\RequirePackage{fix-cm}
\documentclass[smallcondensed]{svjour3}     
\smartqed  
\usepackage{graphicx}
\usepackage{amsmath}
\usepackage{float}
\usepackage[square,sort,comma,numbers]{natbib}
\begin{document}

\title{Proposing a Model for Predicting Passenger Origin-Destination in Online Taxi-Hailing Systems
}

\titlerunning{Predicting Passenger Origin-Destination in Online Taxi-Hailing Systems}        

\author{Pouria Golshanrad  \and
        Hamid Mahini  \and
        Behnam Bahrak 
}


\institute{Pouria Golshanrad \at
	School of Electrical and Computer Engineering, University of Tehran, Tehran, Iran \\
              \email{pouria.golshanrad@ut.ac.ir}           
			\and
           Hamid Mahini \at
           School of Electrical and Computer Engineering, University of Tehran, Tehran, Iran
           \and
           Behnam Bahrak \at
           School of Electrical and Computer Engineering, University of Tehran, Tehran, Iran
}

\date{Received: date / Accepted: date}

\maketitle

\begin{abstract}
	Due to the significance of transportation planning, traffic management, and dispatch optimization, predicting passenger origin-destination has emerged as a crucial requirement for intelligent transportation systems management. In this study, we present a model designed to forecast the origin and destination of travels within a specified time window. To derive meaningful travel flows, we employ K-means clustering in a four-dimensional space with a maximum cluster size constraint for origin and destination zones. Given the large number of clusters, we utilize non-negative matrix factorization to reduce the number of travel clusters. Furthermore, we implement a stacked recurrent neural network model to predict the travel count in each cluster. A comparison of our results with existing models reveals that our proposed model achieves a 5-7\% lower mean absolute percentage error (MAPE) for 1-hour time windows and a 14\% lower MAPE for 30-minute time windows.
\end{abstract}

\section{Introduction}
A significant volume of online ride-sharing data is generated during daily trips, encompassing crucial information about users' travel times, origins, and destinations. This information can be utilized to extract travel patterns for accurate transportation planning and management strategies. Passenger origin-destination (OD) prediction has emerged as a vital component for intelligent transportation systems management. Specifically, this prediction provides data on the number of travels and their origin and destination zones for subsequent time windows, facilitating more efficient driver distribution by adjusting prices. Furthermore, it enables service providers to disable ride-sharing services for requests with a low probability of matching with other requests during travel, particularly if the ride-sharing service price is lower than the standard service price.

In this research paper, we present a perspective on travel clustering utilizing the K-means algorithm in four-dimensional space to identify origin and destination hotspots. Employing two-dimensional clustering on origin and destination points independently may fragment traffic patterns into insignificant segments, consequently diminishing the information contained within these patterns. In essence, due to the disregard of the relationship between origin and destination hotspots in two-dimensional clustering, the resulting outcomes may be of inferior quality. 

In contrast to the previously discussed method, four-dimensional clustering can effectively extract high-quality and meaningful travel flow clusters in smaller quantities by associating origin and destination points during the clustering process. We analyze passenger travel data from Tapsi\footnote{The website of Tapsi is https://tapsi.ir/ (last accessed on June 2, 2023).}, a prominent online taxi-hailing company in the Middle East. The data was sampled from 2018 and 2019. Due to the significant number of extracted clusters, model training necessitates substantial processing resources. Consequently, we employ non-negative matrix factorization (NMF) \cite{cichocki2009fast,fevotte2011algorithms} to reduce the size of the OD vectors. Techniques such as principal component analysis (PCA) do not perform as efficiently as the NMF algorithm, as negative values in the coefficient matrix violate the initial physical meaning and further impair prediction accuracy \cite{li2018hybrid}. The dataset comprises 100 samples of 45 consecutive days from various parts of 2018 and 2019, featuring diverse weather-conditions and holidays. Out of these, 38 days are used for training and 7 days for testing. 

We propose a stacked recurrent neural network model comprising a non-linear supervised feature extraction component and a linear regression component. The non-linear supervised feature extraction component can extract features with a linear relationship to OD vectors for the subsequent time-interval. The linear regression component utilizes the extracted features to estimate OD vectors for the next time-interval. In addition to previous time-interval OD vectors, we incorporate weather and time features into the input vectors to enhance prediction accuracy. 

We evaluate the model's prediction capabilities using simple recurrent cells \cite{connor1994recurrent}, long short-term memory (LSTM) \cite{hochreiter1997long}, and gated recurrent units (GRU) \cite{cho2014learning} as three distinct types of recurrent neural network cells for the non-linear supervised feature extraction component. The GRU cell demonstrates robust prediction capabilities for short time windows. Furthermore, we compare the proposed model with several existing models, including NMF-vector autoregression (NMF-VAR) \cite{li2018hybrid}, multi-layered perceptron (MLP), and support vector machine (SVM) \cite{smola2004tutorial}. Our evaluation reveals that the proposed model achieves a 5-7\% lower MAPE for 1-hour time windows and a 14\% lower MAPE for 30-minute time windows compared to existing models.

\paragraph{In this paper, we present the following key contributions:}
\begin{itemize}
	
	\item {We introduce a novel approach for clustering and preprocessing origin-destination travel data to be utilized in a demand prediction model. This method aims to identify travel patterns and compress the extracted information, making it suitable for input into the prediction model.}
	\item{We develop a neural network-based model that predicts the number of trips between origin-destination zones using historical data, along with additional features such as weather-conditions and temporal characteristics. The improved prediction accuracy compared to existing methods can assist taxi-hailing companies in implementing a more equitable pricing system and enhancing their operational planning.
	}
	
\end{itemize}

The remainder of this paper is structured as follows: Section 2 offers a comprehensive review of the relevant literature. In Section 3, we elucidate three distinct categories of travel flow clustering. Our methodology for origin-destination prediction is delineated in Section 4. Section 5 is dedicated to the presentation and analysis of the experimental results. Lastly, we summarize the principal findings and conclusions in Section 6.

\section{Related Work} 
The prediction of passenger OD in transportation systems has been a topic of interest for researchers and practitioners alike, as it plays a crucial role in transportation planning, traffic management, and service optimization. Various approaches have been proposed in the literature to address this problem, ranging from traditional statistical methods to advanced machine learning techniques. In this section, we provide a comprehensive review of the existing literature on OD prediction, focusing on two main aspects: prediction of travel count and extraction of OD hotspots. We discuss the strengths and limitations of different methods and highlight the gaps in the current state of the art, which our proposed model aims to address.

\subsection{Prediction of Travel Count}
Accurate prediction of travel count between origin and destination hotspots is essential for efficient transportation planning and management. Various methods have been proposed in the literature to estimate the number of travels between predefined zones for specific time-intervals. These methods encompass a wide range of techniques, from traditional statistical models to advanced machine learning algorithms, each with its own set of advantages and limitations. In this subsection, we provide an overview of the most prominent approaches for predicting travel count, discussing their key features, underlying assumptions, and performance in different contexts. This review serves as a foundation for understanding the current state of the art in travel count prediction and identifying the areas where our proposed model can contribute to the existing body of knowledge.

In the study by Toque et al. \cite{toque2016forecasting}, an innovative approach for forecasting dynamic OD matrices in a subway network was proposed using LSTM recurrent neural networks. The effectiveness of this approach was evaluated using a real smart card dataset from the public transport network of Rennes Metropole, France. The results demonstrated that the LSTM model, which incorporated both subway and bus data, outperformed traditional calendar and Vector Autoregression (VAR) models in predicting OD matrices over a 15-minute time horizon. This research highlights the potential of machine learning techniques, particularly LSTM, in improving the accuracy of short-term mobility demand forecasting in public transport networks.

For public transportation like bus and subway, it is clear where the bus stops and subway stations are located. So there is no need for preprocessing to find out the hotspots, it does not include the prediction research works based on the real origin and destination of the travelers. Prediction based on the users' arbitrary origin and destination locations, like online taxi-hailing systems is more challenging. Also, the size of the OD matrix for these types of problems is much larger than the fixed origin-destination problems' OD matrix, thus the model training requires larger processing resources. Toque showed that the LSTM neural network has acceptable prediction power and can predict the number of travels between the origin and destination hotspots for 15-minute time-intervals with a suitable error rate. 

In the study by Ickowicz et al. \cite{Ickowicz2015}, the authors devised a framework for estimating the OD matrix by utilizing ticket records and survey data from public transport users. This approach does not necessitate a prior OD matrix; instead, it models passenger behavior through survey data. The authors proposed a robust version of the estimator to circumvent biases introduced by the survey and constructed a regression estimation procedure that accounts for the influence of exogenous variables, such as weather-conditions and time of the year.

Their method diverges from studies like ours, which require a prior OD matrix and assume knowledge of where people board and alight from public transport. Instead, their method models passengers' behavior on trains and buses in Sydney using survey data and employs diagonalization of the partial OD matrix to reduce the space parameter, ultimately deriving a consistent global OD matrix estimator. The authors develop an estimation procedure based on available ticket records and/or previous surveys, and estimate the whole matrix structure through eigenvectors. Furthermore, they construct a regression estimation procedure that accounts for the influence of exogenous variables, such as weather-conditions and time of the year, providing a more comprehensive understanding of factors affecting public transport usage.

Liu et al. \cite{Liu2019} introduced a Contextualized Spatial-Temporal Network (CSTN) to predict taxi demand between all region pairs in future time-intervals. The model, which incorporates local spatial context, temporal evolution context, and global correlation context, was trained and evaluated on a large-scale dataset of 132 million taxi trips from New York City. The CSTN outperformed other methods in both taxi origin-destination demand prediction and origin demand prediction tasks, achieving an OD-MAPE of 24.93\% and an O-MAPE of 12.92\%. 

The authors divide a city into non-overlapping grid cells using longitude and latitude coordinates. Each rectangular cell corresponds to a distinct geographical region within the city. Their partitioning approach is static and does not take into account the relationship between the origin and destination points of individual trips during the partitioning process.

Li et al. \cite{li2018hybrid} introduced a two-phase hybrid algorithm for estimating OD flows. Initially, they employed the NMF algorithm to tackle the large-scale OD matrix. Subsequently, they utilized the VAR model to estimate the coefficient matrix for the upcoming time window, based on the prior time-series coefficient matrices extracted through the NMF algorithm. However, the VAR model is limited in its ability to capture non-linear relationships between the coefficient matrices.

Reddy and Chakroborty \cite{reddy1998fuzzy} introduced an assignment algorithm grounded in fuzzy inference principles to estimate OD matrices from link volume counts, overcoming the drawbacks of conventional home-interview survey techniques. The algorithm generates the OD elements' relative contributions to link volumes using a flow-dependent approach and employs a recursive method for the OD estimation, resulting in a more cost-effective and precise estimation.

In Ying et al.'s research \cite{ying2016novel}, a new approach for dynamic OD flow estimation was developed, utilizing the polynomial trend model and Kalman filtering theory. The state variables were represented as the differences between the actual OD flow and its historical values, and they were modeled as a stochastic progression with a sliding trend. To estimate and predict the dynamic OD matrix, a polynomial trend filtering model was devised.

Tanaka et al. \cite{tanaka2016estimation} developed a convex quadratic optimization model for estimating OD matrices of bus line passengers from input-output data in public transportation systems. This model addresses the challenges faced by organizations in collecting accurate OD data due to financial constraints. Since the OD matrix of a bus line is small and the locations of bus stops are specific, there is no need for clustering techniques to extract origin and destination hotspots.  

Zuniga et al. \cite{zuniga2021estimation} introduced a dynamic real-time OD matrix prediction methodology based on the MLP neural network for public transportation, utilizing fare systems data. The method incorporates features such as the day of the week, day interval, and calendar information, in addition to aggregated historical travel data from previous time-intervals. However, using historical data in an aggregated format rather than a time series format may negatively impact the seasonality and trend patterns of the data.

In the study by Zhang et al. \cite{zhang2017deep}, the authors developed a deep learning-based method called ST-ResNet to predict the inflow and outflow of crowds in different regions of a city. This model captures spatial and temporal dependencies, as well as external factors such as weather and events. ST-ResNet employs convolution-based residual networks to model spatial properties and three temporal properties: closeness, period, and trend. The approach was tested on datasets from Beijing and New York City. However, the grid-based clustering method may fragment meaningful origin and destination hotspots into multiple zones, resulting in a large number of travel flows. Additionally, the traffic flow between an origin and destination hotspot is transformed into a sequence of inflow and outflow between zones, necessitating route information and leading to a more complex model.

An alternative approach to addressing these problems involves the use of statistical methods. Unlike previous models that were trained on observed OD matrices to estimate the next time-interval OD matrix, statistical models focus on estimating the parameters of the OD matrix distribution. Li et al. \cite{li2005bayesian} proposed a Bayesian statistical model for estimating traffic flows on the links between city zones. In this approach, a path from an origin to a destination consists of a sequence of nodes connected by links. They assumed that the OD matrix follows a multivariate Poisson distribution and employed the expectation-maximization (EM) algorithm to estimate the distribution parameters. As previously mentioned, this type of map zoning approach requires travel route data. Some statistical models assume that each OD flow has an independent Poisson distribution from the other OD flows distribution \cite{spiess1987maximum,vardi1996network,tebaldi1998bayesian,hazelton2000estimation}. However, such an assumption may introduce biases in the estimation.

Wang et al. \cite{wang2021route} developed a comprehensive toolkit for demand estimation and route planning, specifically tailored for long-distance express bus lines in the context of hyperurbanization and jobs-housing imbalance in Beijing. By utilizing mobile phone location data for origin-destination estimation, their proposed method determines the optimal route layout, stop spacing, and operation distance threshold for express bus services in commuter corridors. However, their estimation method does not take into account any additional features, relying solely on the OD distribution derived from mobile location data.

In the research conducted by Kang et al. \cite{kang2020procedure}, a comprehensive algorithm was developed to generate OD matrices and passenger load profiles for public transit networks utilizing smart card transaction data. The method was effectively applied to the Tehran bus rapid transit (BRT) network, identifying 52\% of transactions as part of a trip and analyzing transfers between lines. The authors presented an integrated procedure designed to generate OD matrices for public transportation fare collection systems that do not have transaction data available at the destination stop.

In the study by Zwick et al. \cite{zwick2022ride}, the authors investigated pre-pandemic demand data from MOIA's ride-pooling services in Hamburg and Hanover, using spatial and random forest regressions to comprehend the spatial characteristics of ride-pooling trip origins in both cities. Additionally, they explored the generalizability of their findings from one city to another, facilitating spatial predictions beyond areas with existing services. Our contribution extends this work by integrating additional external features, such as weather-conditions and time characteristics, and employing a novel neural network-based model for predicting passenger origin-destination. As a result, our proposed model demonstrates improved prediction accuracy compared to existing methods.

In their study, Ge et al. \cite{ge2021review} developed a model for predicting passenger OD in online taxi-hailing systems by employing K-means clustering and NMF. They also introduced an approach for clustering and preprocessing OD travel data, which substantially enhanced prediction accuracy. In contrast, our contribution focuses on a more advanced clustering technique that does not require specifying the number of clusters as input and incorporates additional external features, resulting in a more comprehensive understanding of passenger behavior.

In the book "Public Transport Planning with Smart Card Data" (2017), edited by Fumitaka Kurauchi and Jan-Dirk Schmöcker \cite{kurauchi2017public}, the authors delve into the potential of utilizing smart card data for public transport planning and evaluation. They discuss a variety of methods for estimating OD matrices, routes, and activities based on smart card data. However, the methods proposed in the book do not take into account additional features such as weather-conditions and temporal characteristics, which could enhance prediction accuracy. Moreover, the techniques presented in the book might not be sophisticated enough to extract intricate features from the data.

\subsection{OD Hotspots Extraction}
The extraction of OD hotspots is a critical aspect of predicting travel demand in transportation systems. Identifying these hotspots enables the representation of complex travel patterns in a more compact and meaningful manner, which is essential for the effective functioning of prediction models. Various techniques have been proposed in the literature for preprocessing and aggregating origin and destination points, each with its own set of advantages and challenges. In this subsection, we review the most prominent approaches for OD hotspots extraction, discussing their underlying principles, applicability in different contexts, and potential impact on the overall prediction accuracy. 

Yue et al. \cite{yue2009mining} employed a clustering approach to group spatiotemporally similar pick-up and drop-off points, aiming to uncover meaningful patterns in time-dependent attractive areas and movement patterns from taxi trajectory data. By considering variations in people's interests throughout different times of the day, week, and season, the authors were able to identify areas with high levels of attractiveness and their associated travel behaviors. 

Mungthanya et al. \cite{Mungthanya2019} introduce a method for utilizing taxi trajectory data to create a taxi OD matrix that dynamically adapts in both spatial and temporal dimensions. This approach allows for flexible origin and destination zone sizes and locations, enabling the matrix dimensions to change as needed. The authors also develop a novel similarity measure for OD matrices, which uncovers the time periods with the highest and lowest taxi travel demand, as well as the periods exhibiting the most and least consistent taxi travel demand patterns. In the context of our problem, the large number of travel flows necessitates the incorporation of time as a fifth dimension, leading to an increased number of clusters containing fewer data points and random patterns.

Clusters with a small number of members may exhibit random behavior, making accurate predictions challenging. Jahnke et al. \cite{jahnke2017identifying} suggested using the density-based spatial clustering of applications with noise (DBSCAN) \cite{ester1996density} algorithm to identify high-intensity origin and destination hotspots throughout the city, aiding in the visual analysis of travel behavior. Utilizing two-dimensional clustering to pair origin and destination zones results in an enlarged OD matrix, reducing the significance and quality of OD flows. However, four-dimensional clustering produces fewer travel flow clusters by considering both the origin and destination of trips during the clustering process.

Fiori et al. \cite{fiori2016decoclu} introduced a new approach, called DeCoClu, for extracting the topology of a public transport network by employing a consensus clustering density-based method. This technique effectively determines the geographical locations of stops and outlines the logical pathways of routes by analyzing static information from time series of positioning signals, such as GPS data. Experimental results on real-data sets showcase the effectiveness of DeCoClu in accurately identifying route stops and enhancing the overall comprehension of the public transport network topology.

In the research conducted by Viggiano et al. \cite{viggiano2017journey}, they developed a journey-based approach to analyze multi-modal public transportation networks using smart card data. The primary objective of their study was to comprehend the roles of various public transport modes within the network. To achieve this, they employed a methodology that involved clustering stops and stations into zones and categorizing OD pairs based on the mode or combination of modes utilized during the journey. 

He et al. \cite{he2021space} developed a method to categorize public transit smart card users' spatiotemporal behaviors using dynamic time warping, hierarchical clustering, and a sampling technique, aiming to improve public transit services by understanding users' travel patterns and preferences. The authors demonstrated the method's effectiveness through visualizations of daily trajectories and space-time path plots. However, the study faced limitations in computation time, exclusion of dissimilarity factors like travel purpose, and potential inaccuracies in destination estimations, which could affect clustering results.

\section{OD Hotspots}
In contemporary online taxi-hailing systems, users have the ability to request taxis for transportation between two arbitrary points on a map. Consequently, unlike public transportation methods such as buses or trains, the origin and destination hotspots are not immediately apparent. Therefore, it is necessary to conduct preprocessing in order to identify pick-up and drop-off hotspots, as well as to divide the city map into zones based on these hotspots. The three most commonly employed preprocessing techniques for hotspot extraction include:

\subsection{Grid-Based Clustering}
\begin{figure}
	\includegraphics[width=0.7\textwidth]{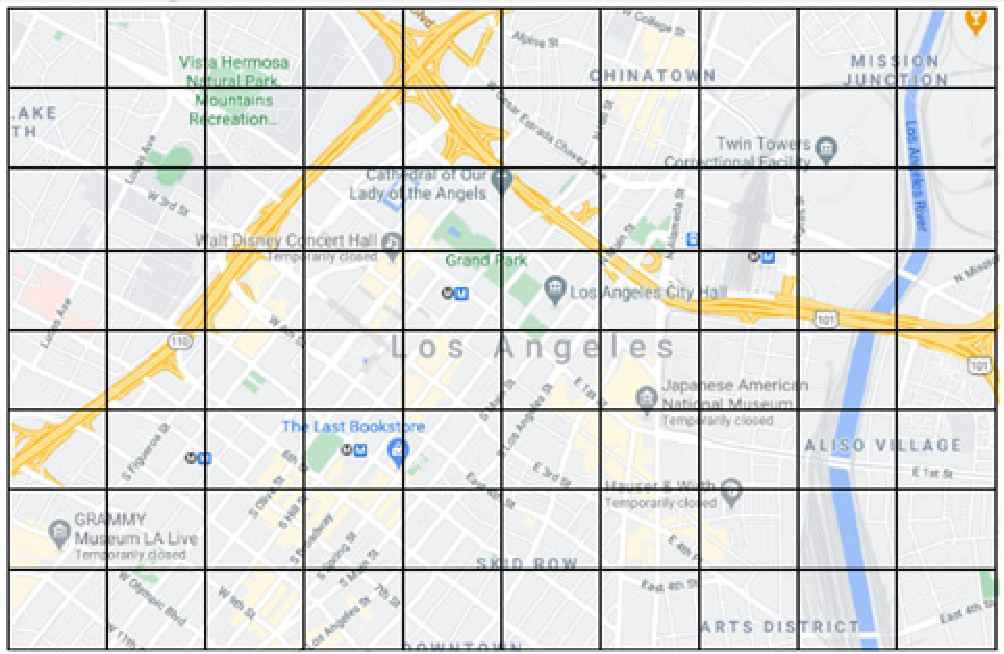}
	\caption{The most straightforward approach to clustering origins and destinations is by employing grid-based clustering techniques.}
	\label{fig:grid}
\end{figure}

\begin{figure}
	\includegraphics[width=0.5\textwidth]{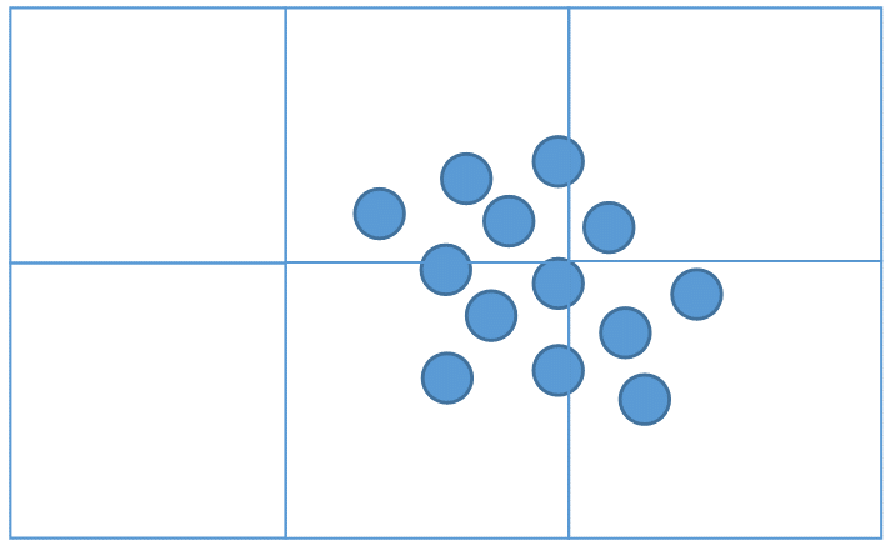}
	\caption{The primary issue in grid-based clustering is the diminished cluster quality resulting from the division of hotspots between segments.}
	\label{fig:grid_problem}
\end{figure}

A straightforward and efficient approach for clustering origin and destination points is employed by Zhang et al. \cite{zhang2012mining} for semantic mining of taxi flows, and Calabrese et al. \cite{calabrese2011estimating} for estimating the OD flows using users' mobile phone location data. This method partitions the city map into equal squares, as illustrated in Fig. \ref{fig:grid}. The primary issue with this method is that a hotspot, such as a hospital, subway station, or shopping center, may be divided among multiple squares, as demonstrated in Fig. \ref{fig:grid_problem}. This problem causes origin and destination hotspots to lose their semantic meaning. Another limitation of grid-based clustering is that after clustering, the Cartesian product of zones must be utilized to generate a single OD matrix, resulting in numerous OD matrix entries with no travel demand.


\subsection{Two-Dimensional Clustering}
Various common approaches, such as K-means, agglomerative hierarchical clustering, and DBSCAN, are employed for detecting hotspots of taxi trip origins and destinations in two-dimensional map space \cite{jahnke2017identifying}. For instance, Tang et al. \cite{tang2015uncovering} utilized the DBSCAN clustering algorithm to cluster pick-up and drop-off locations, which facilitated the generation of an OD matrix for traffic distribution estimation. In this study, we employ the K-means clustering algorithm, an optimization-based clustering method \cite{arthur2007k}, which aims to enhance intra-cluster similarity by minimizing the value provided in Eq. (\ref{eqs:kmeans_cost}). The K-means cost value $(J)$ is calculated by summing the squared Euclidean distances between each data point $(\vec{x_{q}})$ and its corresponding cluster center $(\vec{\mu_{i}})$, where $C$ is the number of clusters and $w_{i}$ represents the set of points belonging to the $i$-th cluster. To decrease the K-means cost value $(J)$, the origin and destination vectors $(\vec{x_{q}})$ should converge to their respective cluster center vectors $(\vec{\mu_{i}})$. According to Eq. (\ref{eqs:scatter}), by reducing the K-means cost, the inter-cluster scatter $(S_{W})$ decreases, and consequently, the intra-cluster scatter $(S_{B})$ increases due to the constant value of the total points scatter $(S_{T})$. In each iteration of the K-means algorithm, points are assigned to the nearest clusters.

\begin{equation} 
\label{eqs:kmeans_cost}
J = \sum_{i=1}^{C}\sum_{x_{q}\in w_{i}}^{ }||\vec{x_{q}}- \vec{\mu_{i}}||^{2}
\end{equation}
\begin{equation} 
\label{eqs:scatter}
S_{T} = S_{W} + S_{B}
\end{equation}

In order to ensure the informativeness and practicality of predictions for large zones, it is essential to impose a maximum size limitation on the origin and destination hotspots. This limitation is contingent upon the sparsity and quality of the data, which may vary across different datasets. While smaller hotspots are preferable, dividing the data into numerous small hotspots may result in stochastic behavior. To determine the optimal maximum size limitation, we employ the K-means clustering algorithm with varying numbers of seeds on origin and destination points, aiming to identify the approximate number of clusters with suitable maximum intra-distance. Further details regarding these results can be found in Subsection \ref{four_dimensional_kmeans_clustering}.

\begin{figure}
	\includegraphics[width=0.7\textwidth]{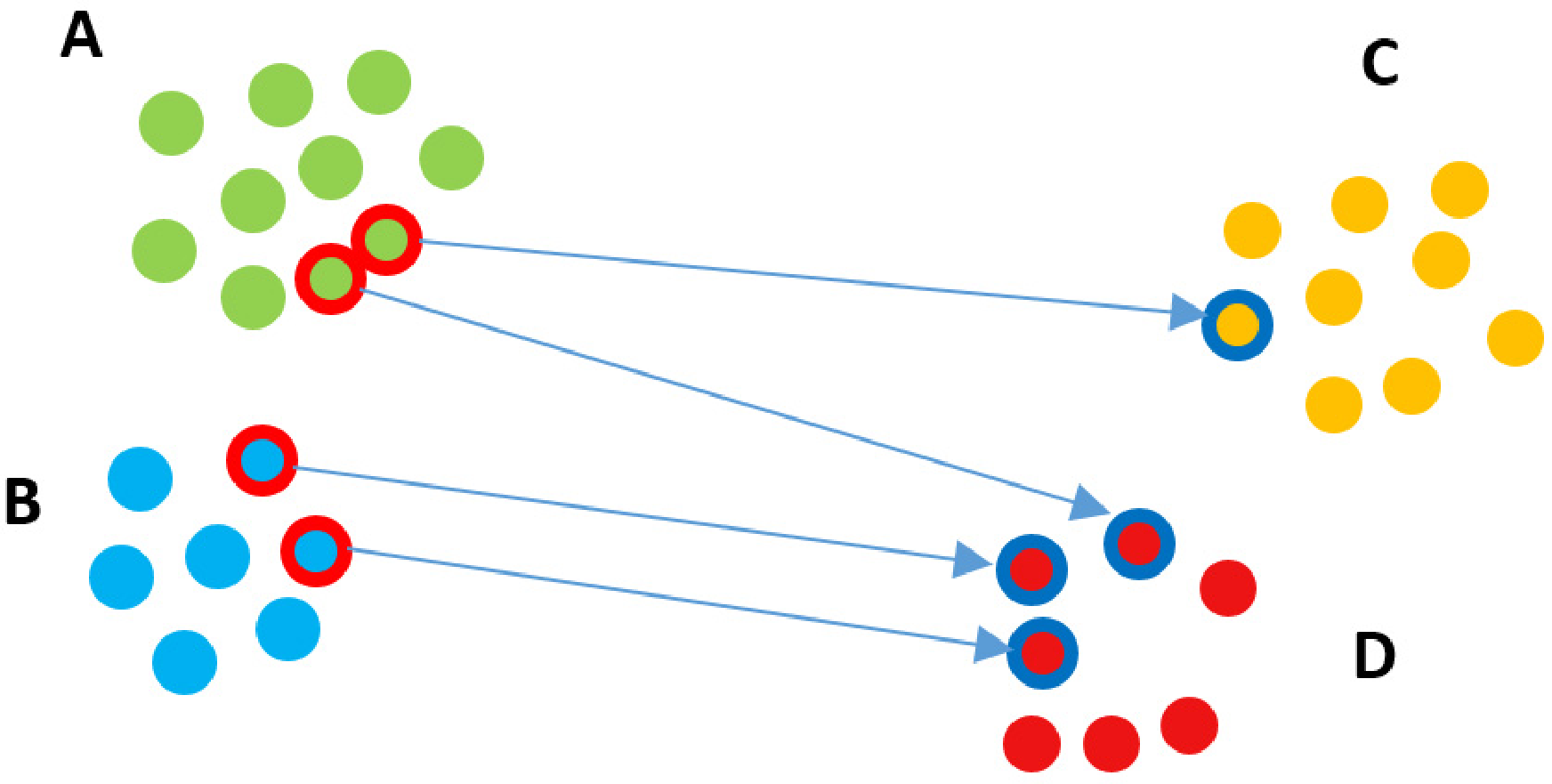}
	\caption{Insufficient extraction of travel patterns using two-dimensional clustering techniques.}
	\label{fig:2D_clustering}
\end{figure}

Following the extraction of origin and destination hotspots through the implementation of the K-means clustering algorithm, it is necessary to associate the corresponding origin and destination hotspots. However, certain pairs of origin and destination hotspots may exhibit no travel demand, resulting in a potentially reduced OD matrix. Moreover, the utilization of two-dimensional clustering for separate origin and destination points may adversely impact the quality of the identified travel patterns.
As illustrated in Figure \ref{fig:2D_clustering}, clusters $A$ and $B$ represent two origin hotspots, while clusters $C$ and $D$ denote two destination hotspots. Based on this clustering, the OD matrix comprises four entries, with three of them containing a non-zero number of travels. Due to the distinct clustering of origin and destination points, a single travel pattern has been divided into three OD matrix entries. If the clustering algorithm had grouped the points with the thick margin as one origin and one destination hotspot, the travel pattern would have been represented as a single OD matrix entry.

\subsection{Four-Dimensional Clustering}
\begin{figure}
	\includegraphics[width=0.7\textwidth]{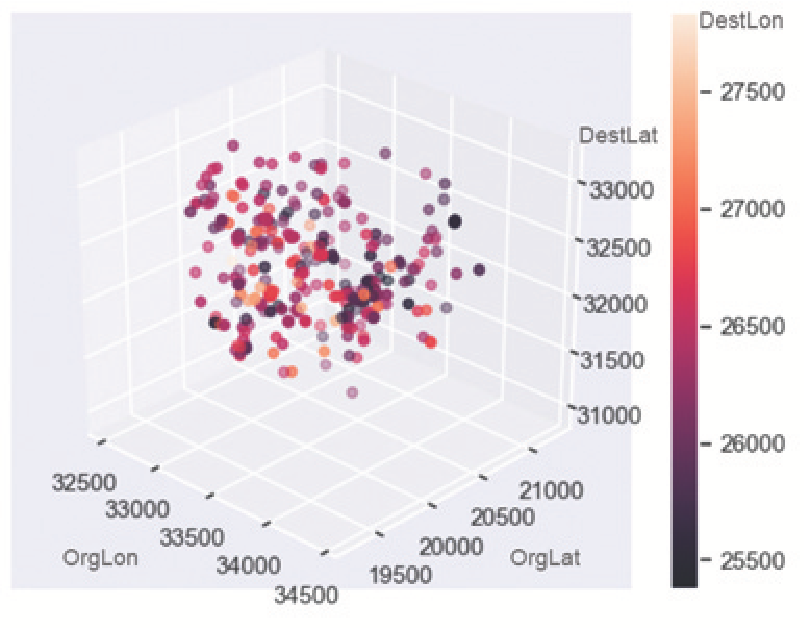}
	\caption{A travel cluster represented in four-dimensional space, consisting of origin latitude $(Org_{Lat})$, destination latitude $(Dest_{Lat})$, origin longitude $(Org_{Lon})$, and destination longitude $(Dest_{Lon})$.} 
	\label{fig:4D_clustering}
\end{figure}

As discussed in the preceding subsection, it has been observed that two-dimensional clustering leads to an enlarged OD matrix, which adversely impacts travel patterns. An alternative approach is to transition into a four-dimensional space, where the dimensions are constituted by the latitude and longitude of both origin and destination points.
Each point $(Org_{Lat}, Dest_{Lat}, Org_{Lon}, Dest_{Lon})$ within this space represents a trip on a two-dimensional map. Proximity between points in this space corresponds to trips with origins and destinations in close proximity to one another. The K-means clustering algorithm is employed for travel points in this four-dimensional space, and the algorithm is executed with varying numbers of clusters to identify the optimal clustering with minimal intra-cluster maximum distance mean and a reduced number of clusters.
Figure \ref{fig:4D_clustering} illustrates one of the travel clusters. In order to visualize travel points in the four-dimensional space, a color range is utilized as the fourth dimension within a three-dimensional space. Point coordinates are determined based on the distance (in meters) from an arbitrary origin.

\begin{figure}
	\includegraphics[width=0.6\textwidth]{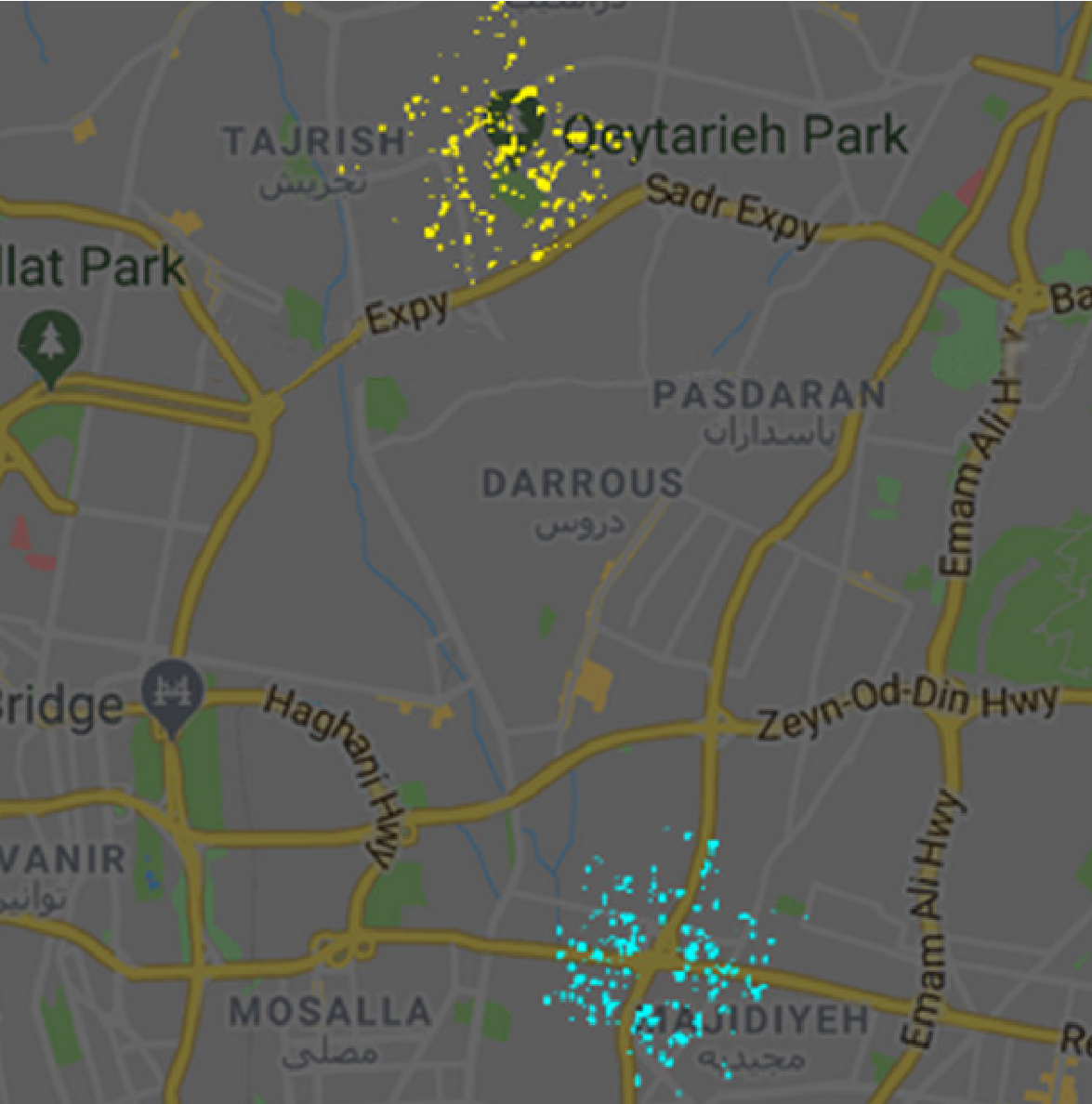}
	\caption{A travel cluster in a two-dimensional space, where the lower points represent travel origins and the upper points denote travel destinations. }
	\label{fig:4D_clustering_map}
\end{figure}

This methodology presents two primary benefits. Firstly, it reduces the size of the OD matrix. In a four-dimensional space, origin and destination hotspots are not paired, and each travel cluster possesses its own distinct hotspots. Consequently, an OD vector is employed instead of an OD matrix. Secondly, the resulting clusters exhibit high quality, providing more insightful information regarding travel patterns. This is due to the simultaneous participation of both origin and destination points in travel clustering and the distribution of points across fewer entries. To determine the optimal maximum size limit for origin and destination hotspots, the clustering algorithm is executed with varying numbers of clusters, taking into account data maturity and domain expert opinions. The travel cluster depicted in Fig. \ref{fig:4D_clustering} is illustrated on the map in Fig. \ref{fig:4D_clustering_map}.

\section{Predicting Travel Requests}
In developing a travel request prediction model, two potential methodologies can be considered: (1) constructing a unified model for all travel clusters, and (2) devising distinct models for each travel cluster, contingent upon their respective travel patterns. For instance, a time series neural network may be suitable for predicting one travel cluster, while a straightforward statistical model could suffice for another. It is crucial to acknowledge that managing a considerable number of clusters is necessary for both strategies. In this paper, our primary focus lies on the first approach.

\subsection{Decomposition of OD Vectors}
As discussed in the preceding section, the application of the K-means clustering algorithm on travel points within a four-dimensional space resulted in numerous travel clusters, each with a maximum intra-distance limitation for origins and destinations. When training models such as neural networks, a large input vector size leads to an increased number of trainable variables. Decomposition techniques, such as PCA, can be employed to reduce the size of the origin-destination vector.
PCA diminishes the dimensionality of the OD vector by selecting eigenvectors with significantly larger eigenvalues \cite{jolliffe2011principal}. As previously mentioned, several studies have employed PCA for OD matrix decomposition; however, a major issue with utilizing PCA in this context is that the negative values produced by PCA may not accurately represent the real-world situation, consequently affecting the prediction accuracy \cite{li2018hybrid,djukic2012efficient,berry2007algorithms}.
Numerous studies have demonstrated that the NMF algorithm ensures the extraction of integrated features from data through non-negative values during the analysis process \cite{lee1999learning,berry2007algorithms,devarajan2008nonnegative}. The NMF algorithm has been widely employed in various domains, including spectral data analysis, facial feature learning, and semantic text feature extraction \cite{lee1999learning,berry2007algorithms,devarajan2008nonnegative}. Additionally, it has been utilized in DNA gene expression analysis \cite{brunet2004metagenes} and protein interaction clustering \cite{greene2008ensemble}.
Utilizing the NMF algorithm as delineated in \cite{cichocki2009fast,fevotte2011algorithms}, each OD vector is decomposed into a coefficient vector and a constant basis matrix, as illustrated in Eq. (\ref{eqs:NMF}). The coefficient vector possesses significantly fewer dimensions. The time series coefficient vector is predicted, and subsequently, the basis matrix is employed to inversely transform the predicted coefficient vector into an OD vector. Assuming the total time duration of the data is $(T)$ and each OD vector comprises $(N)$ travel demand clusters within a specific time-interval, the NMF algorithm yields the coefficient vectors in $(M)$ dimensional space, which is considerably smaller than the $(N)$ dimensional space.

\begin{equation} 
\label{eqs:NMF}
\resizebox{0.8\hsize}{!}{$
		\begin{bmatrix}
		v_{1}^{1} & v_{2}^{1}  & \dots  & v_{N}^{1} \\
		v_{1}^{2} & v_{2}^{2}  & \dots  & v_{N}^{2} \\
		\vdots & \vdots & \ddots  & \vdots \\
		v_{1}^{T} & v_{2}^{T}  & \dots  & v_{N}^{T}
		\end{bmatrix}
		=
		\begin{bmatrix}
		c_{1}^{1} & c_{2}^{1}  & \dots  & c_{M}^{1} \\
		c_{1}^{2} & c_{2}^{2}  & \dots  & c_{M}^{2} \\
		\vdots & \vdots &  \ddots & \vdots \\
		c_{1}^{T} & c_{2}^{T}  & \dots  & c_{M}^{T}
		\end{bmatrix}
		*
		\begin{bmatrix}
		b_{11} & b_{12}  & \dots  & b_{1N} \\
		b_{21} & b_{22}  & \dots  & b_{2N} \\
		\vdots & \vdots  & \ddots & \vdots \\
		b_{M1} & b_{M2}  & \dots  & b_{MN}
		\end{bmatrix}
	$}
\end{equation}

\begin{center}
	\resizebox{0.15\hsize}{!}{
		$ N>>M$
	}
\end{center}
NMF is an optimization-based algorithm. The objective is to decompose a given matrix $(V)$ into two matrices $(C, B)$, as shown in Equation (\ref{VCB}), with the constraint that $C \geq 0$ and $B \geq 0$ (Equation \ref{cost}). To achieve this decomposition, the NMF cost function must be minimized (Equation \ref{cost2}). In the context of data dimensionality reduction, the value of $(M)$ is typically selected such that $TM + MN \ll TN$ \cite{fevotte2011algorithms}.

\begin{equation} 
\label{VCB}
V \approx CB
\end{equation}
\begin{equation} 
\label{cost}
\underset{C,B}{min}\; D(V|CB) \; subject\;  \; to\;  C \geq 0 \; , \; B \geq 0
\end{equation}
\begin{equation} 
\label{cost2}
D(V|CB) = \sum_{t=1}^{T}\sum_{n=1}^{N} d_{\beta}([V]_{tn} \;|\; [CB]_{tn})
\end{equation}

The scalar cost function $d_{\beta}(x|y)$, referred to as $\beta$-divergence, is characterized by a single minimum when $x = y$ (\ref{beta_divergance}) \cite{basu1998robust}. This family of cost functions is parameterized by a shape parameter $\beta$, which encompasses the Euclidean distance, Kullback-Leibler (KL) divergence, and Itakura-Saito (IS) divergence as special cases for distinct values of $\beta$ \cite{fevotte2011algorithms}. Specifically, the $\beta$-divergence behaves like IS and KL divergences for $\beta = 0$ and $\beta = 1$, respectively.

\begin{equation} 
\resizebox{0.9\hsize}{!}{$
	\label{beta_divergance}
	d_{\beta}(x | y) = 
	\begin{cases}
	\dfrac{1}{\beta(\beta-1)}(x^{\beta}+(\beta-1)y^{\beta}-\beta x y^{\beta - 1}) &\quad \beta \in \Re,\beta \neq 0,1\\
	x log \frac{x}{y} - x + y &\quad \beta = 1 \\
	\frac{x}{y} - log \frac{x}{y} - 1 &\quad \beta = 0\\
	\end{cases}$}
\end{equation}
\begin{flushleft}
	 The heuristic update rules for the basis and coefficient matrices can be expressed as follows, according to \cite{fevotte2011algorithms}:
\end{flushleft}
\begin{equation} 
B \leftarrow B.\dfrac{C^{T}[(CB)^{.\beta -2}.V]}{C^{T}[CB]^{.(\beta - 1)}}
\end{equation}
\begin{equation} 
C \leftarrow C.\dfrac{[(CB)^{.\beta -2}.V]B^{T}}{[CB]^{.(\beta - 1)}B^{T}}
\end{equation}

\subsection{Prediction Model}
\label{prediction_model_section}
\begin{figure}
	\centering 
	\includegraphics[width=\textwidth]{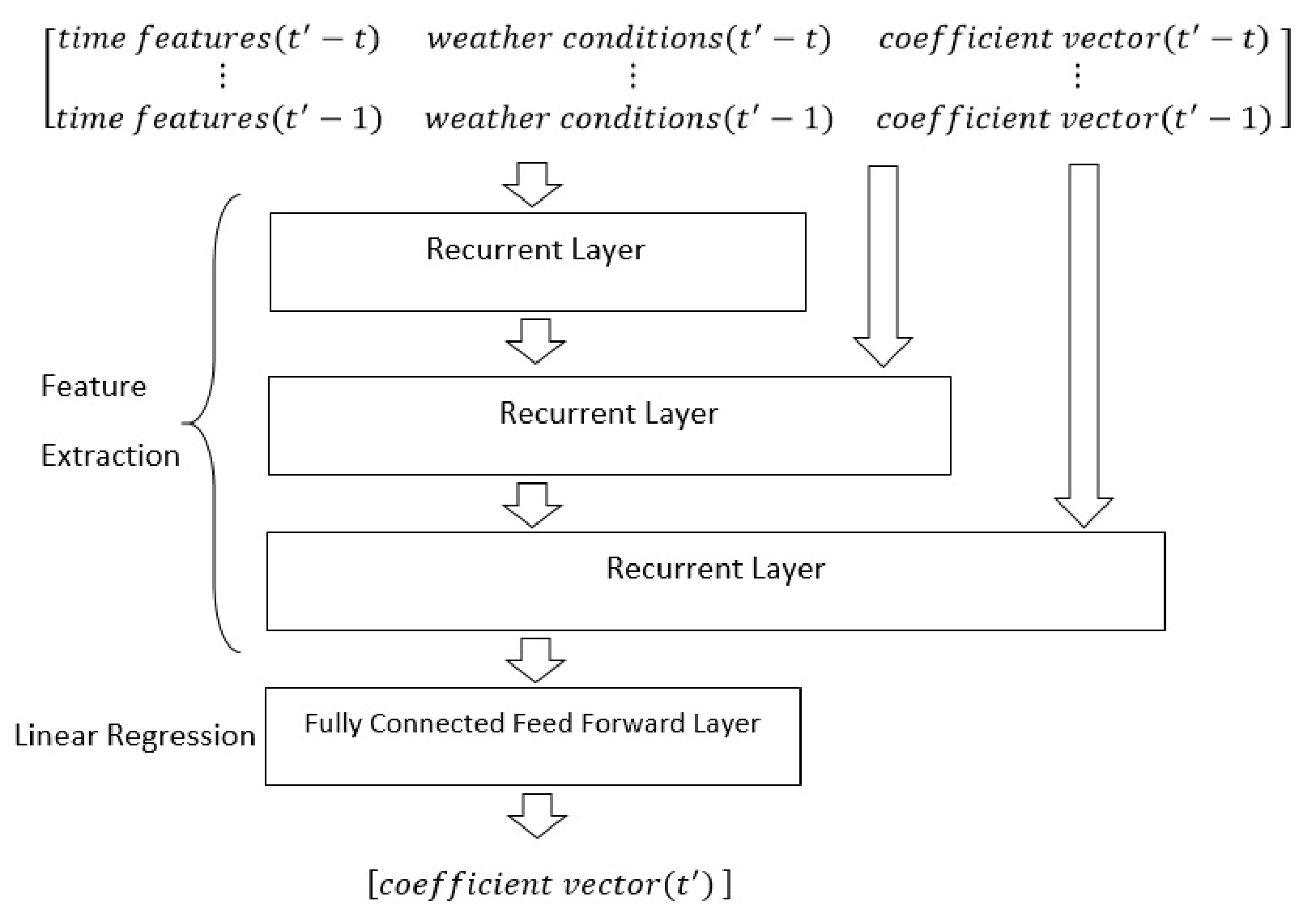}
	\caption{Structure of the NMF-Stacked RNN Model}
	\label{fig:PredictionModel}
\end{figure}

After employing the NMF algorithm to reduce the dimensions of time series OD vectors, we utilize a stacked Recurrent Neural Network (RNN) for feature extraction and a feed-forward fully connected layer for linear regression to predict the coefficient vector of the subsequent time-interval $(t')$, as illustrated in Fig. \ref{fig:PredictionModel}. To obtain the OD vector for the next time-interval, the estimated coefficient vector must be multiplied by the basis matrix acquired from the NMF algorithm. In order to enhance the prediction accuracy of the model, various weather-condition features, such as temperature, pressure, humidity, wind speed, wind direction, dew point temperature, and visibility, are incorporated into the travel clusters data. Additionally, time features like the hour of the day, the day of the week, part of the hour (for time windows shorter than an hour), and is-holiday are employed for travel prediction. Table \ref{tab:features_details} provides further details about the features and their integration into the prediction model. The supervised feature extraction component of the model aids in extracting the most influential features on the output through sequential nonlinear mappings.

\begin{table}[]
	\caption{Adapted features and their expressions for the prediction model}\label{tab:features_details}
	\begin{tabular}{|l|l|}
		\hline
		\textbf{Time Frame Features} & \textbf{Descriptions}                                                                                                                                                                                                                                                                                                   \\ \hline
		Coefficient Vector                & \begin{tabular}[c]{@{}l@{}}This vector is obtained using NMF, representing the \\ compressed encoding of travel counts for a specific \\ time frame.\end{tabular}                                                                                                                                                      \\ \hline
		Hour of the Day                   & \begin{tabular}[c]{@{}l@{}}A 24-dimensional one-hot encoded vector, where \\ each dimension corresponds to an hour of the day.\end{tabular}                                                                                                                                                                                   \\ \hline
		Day of the Week                   & \begin{tabular}[c]{@{}l@{}}A 7-dimensional one-hot encoded vector, where \\ each dimension corresponds to a day of the week.\end{tabular}                                                                                                                                                                                  \\ \hline
		Part of the Hour                  & \begin{tabular}[c]{@{}l@{}}For a 30-minute time frame, this 2-dimensional \\ one-hot encoded vector represents each half of \\ the current hour.\end{tabular}                                                                                                                     \\ \hline
		Is Holiday                        & A binary feature indicating whether the day is a holiday or not.                                                                                                                                                                                                                                                           \\ \hline
		Temperature                       & \begin{tabular}[c]{@{}l@{}}The z-score of temperature based on the previous \\ year's temperature distribution.\end{tabular}                                                                                                                                                                                             \\ \hline
		Pressure                          & \begin{tabular}[c]{@{}l@{}}The z-score of pressure based on the previous \\ year's pressure distribution.\end{tabular}                                                                                                                                                                                                   \\ \hline
		Humidity                          & \begin{tabular}[c]{@{}l@{}}The z-score of humidity based on the previous \\ year's humidity distribution.\end{tabular}                                                                                                                                                                                                   \\ \hline
		Wind Speed                        & \begin{tabular}[c]{@{}l@{}}The z-score of wind speed based on the previous \\ year's wind speed distribution.\end{tabular}                                                                                                                                                                                                       \\ \hline
		Wind Direction                    & \begin{tabular}[c]{@{}l@{}}An 8-dimensional one-hot encoded vector representing \\ one of the following directions: North, North East, \\ North West, East, West, South, South West, \\ South East.\end{tabular}                                                                              \\ \hline
		Dew Point Temperature             & \begin{tabular}[c]{@{}l@{}}The z-score of dew point temperature based on the \\ previous year's dew point temperature distribution.\end{tabular}                                                                                                                                                                         \\ \hline
		Visibility                        & \begin{tabular}[c]{@{}l@{}}A 7-dimensional one-hot encoded vector representing \\ one of the following visibility ranges (in kilometers):\\ Clear (23 km), Thin Haze (3.2 km), Light Haze \\(2.8 km), Moderate Haze (1.9 km), Light Fog (1.3 km), \\ Moderate Fog  (1.1 km).\end{tabular} \\ \hline
	\end{tabular}
\end{table}

\begin{figure}
	\includegraphics[width=0.7\textwidth]{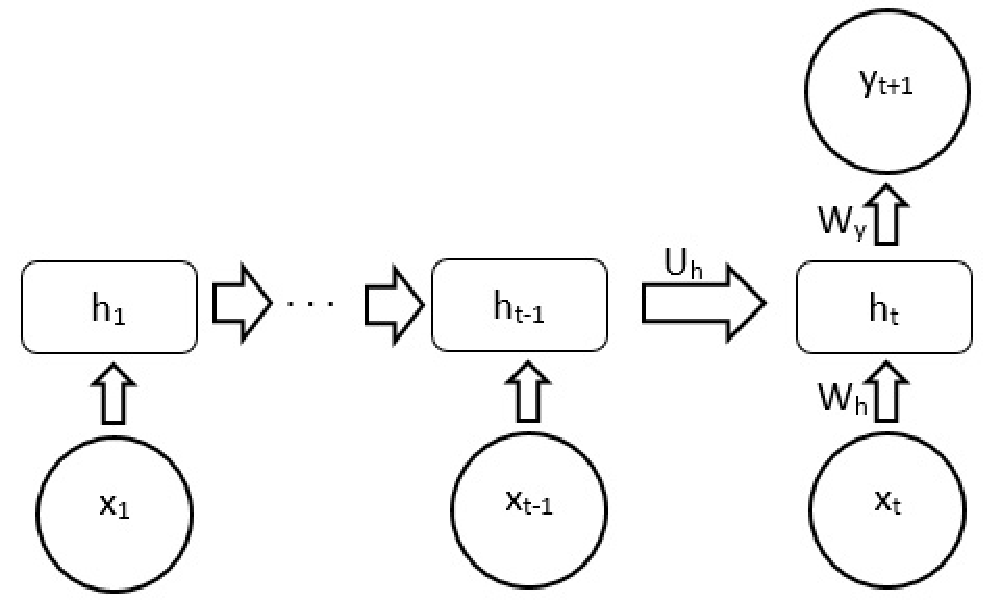}
	\caption{An Unfolded Simple Recurrent Cell}
	\label{fig:RNN}
\end{figure}

The input for our model consists of a sequence of features $(x)$ from the last $(t)$ time-intervals, which includes the coefficient vectors of the previous $(t)$ time-intervals as well as time and weather-condition features. We evaluate three types of recurrent neural network cells in the proposed model: 1) Simple Recurrent, 2) LSTM, and 3) GRU. As illustrated in Fig. \ref{fig:PredictionModel}, the prediction model employs three recurrent layers. An unfolded simple recurrent cell for the input sequence $(x)$ with length $(t)$ is depicted in Fig. \ref{fig:RNN}. In accordance with Eq. (\ref{eq1:RNN}) and Eq. (\ref{eq2:RNN}), a nonlinear mapping is modeled by the $sigmoid$ activation function (\ref{sigmoid}) for each sequence entry, facilitating the discovery of nonlinear relationships among input features. Each state of the cell $(h_t)$ is computed based on the previous state vector value $(h_{t-1})$ and the current time-interval feature vector $(x_t)$. However, this type of recurrent cell exhibits a limitation in that it tends to forget the initial entries of a lengthy input sequence \cite{hochreiter1997long}.

\begin{equation} 
\label{eq1:RNN}
h_{t} = \sigma(W_{h}x_{t} + U_{h}h_{t-1} + b_{h})
\end{equation}
\begin{equation} 
\label{eq2:RNN}
y_{t+1} = \sigma(W_{y}h_{t}+b_{y})
\end{equation}
\begin{equation} 
\label{sigmoid}
\sigma(x) = \frac{1}{1+e^{-x}}
\end{equation}

Hochreiter and Schmidhuber \cite{hochreiter1997long} introduced a recurrent neural network cell, known as LSTM, designed to address the challenges of exploding and vanishing gradients. The LSTM cell incorporates multiple nonlinear mappings utilizing $sigmoid$ and $tanh$ activation functions. Moreover, it is capable of learning when to forget $(f)$ as shown in Eq. (\ref{forget}), when to incorporate input information $(i)$ as depicted in Eq. (\ref{lstm_input}) into the memory cell $(c)$ as illustrated in Eq. (\ref{lstm_memory}), and how to regulate the output gate $(o)$ as demonstrated in Eq. (\ref{lstm_output}). The final hidden state vector $(h_{t+1})$ is presented in Eq. (\ref{lstm_hidden1}) and serves as the output of the LSTM unit, which is denoted as $(y)$ in Eq. (\ref{lstm_hidden2}). It is important to note that the number of trainable variables in an LSTM cell is significantly higher than that of a simple recurrent cell, potentially leading to longer training times and, in some cases, overfitting.

\begin{equation} 
\label{forget}
f_{t} = \sigma(W_{f}x_{t-1} + U_{f}h_{t-1} + b_{f})
\end{equation}
\begin{equation} 
\label{lstm_input}
i_{t} = \sigma(W_{i}x_{t-1} + U_{i}h_{t-1} + b_{i})
\end{equation}
\begin{equation} 
\label{lstm_memory}
c_{t} = f_{t}c_{t-1} + i_{t}tanh(W_{c}x_{t-1} + U_{c}h_{t-1} + b_{c})
\end{equation}
\begin{equation} 
\label{lstm_output}
o_{t} = \sigma(W_{o}x_{t-1} + U_{o}h_{t-1} + b_{o})
\end{equation}
\begin{equation} 
\label{lstm_hidden1}
h_{t} = o_{t}tanh(c_{t})
\end{equation}
\begin{equation} 
\label{lstm_hidden2}
y_{t+1} = h_{t+1}
\end{equation}
\begin{figure}
	\includegraphics[width=\textwidth]{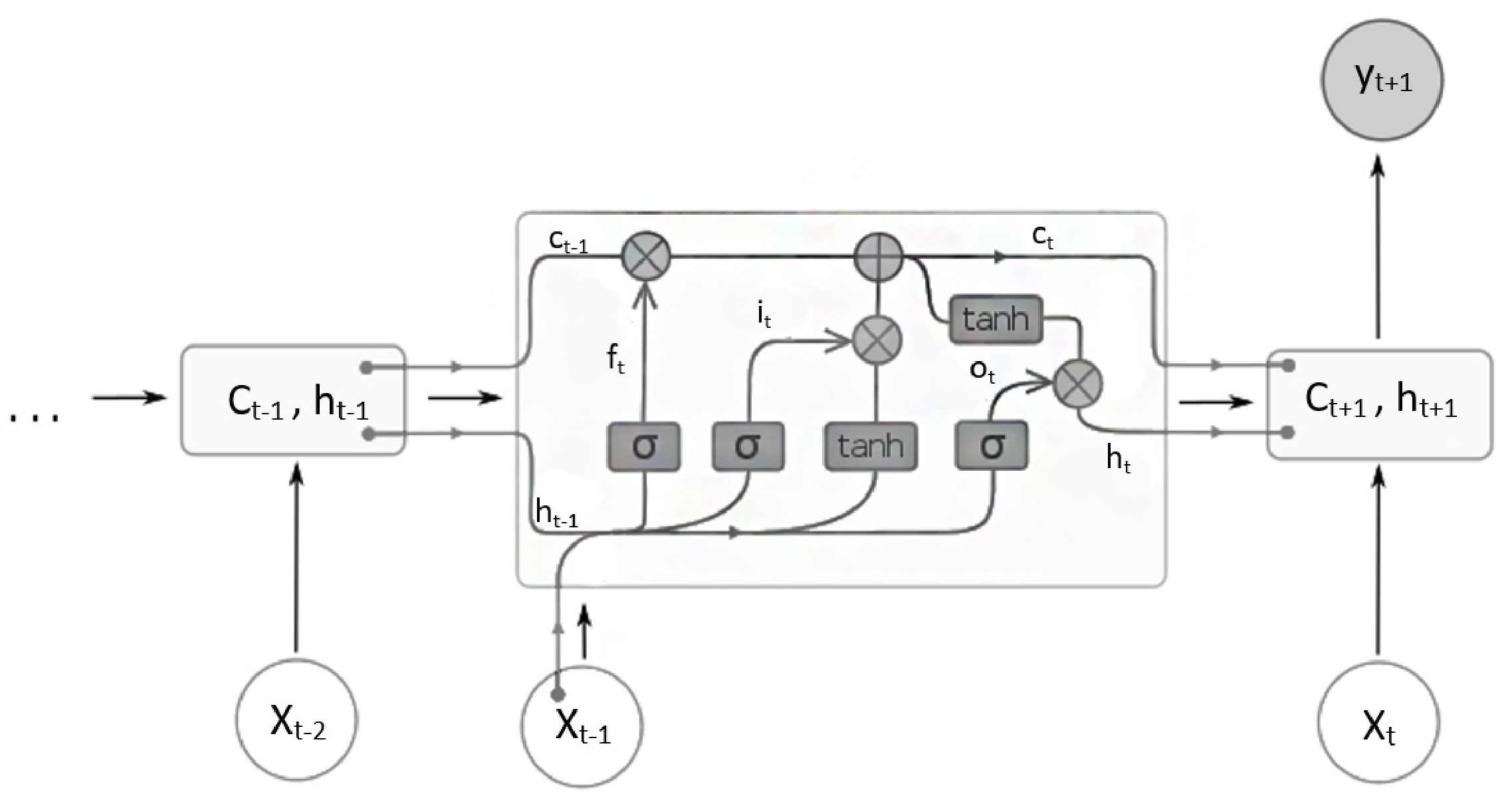}
	\caption{An Unfolded LSTM Cell}
	\label{fig:LSTM}
\end{figure}

Cho et al. \cite{cho2014learning} introduced a gating mechanism for recurrent neural network cells, similar to LSTM, but with only update $(z)$ Eq. (\ref{GRU_update}) and reset $(r)$ Eq. (\ref{GRU_reset}) gates, resulting in fewer trainable variables. The subsequent cell state $(h_t)$ Eq. (\ref{GRU_output}) is determined based on the trained reset and update gates, the feature vector of the previous time window $(x_{t-1})$, and its hidden state $(h_t)$. The final hidden state vector of the GRU unit $(h_{t+1})$ Eq. (\ref{GRU_hidden}) serves as the output of the memory cell ${(y_{t+1})}$. Chung et al. \cite{chung2014empirical} demonstrated that GRU outperforms LSTM on smaller datasets where pattern extraction is more challenging.

\begin{equation} 
\label{GRU_update}
z_{t} = \sigma(W_{z}x_{t-1} + U_{z}h_{t-1} + b_{z})
\end{equation}
\begin{equation} 
\label{GRU_reset}
r_{t} = \sigma(W_{r}x_{t-1} + U_{r}h_{t-1} + b_{r})
\end{equation}
\begin{equation} 
\label{GRU_output}
h_{t} = (1-z_{t})h_{t-1} + z_{t}tanh(W_{h}x_{t-1} + U_{h}(r_{t}h_{t-1}) + b_{h})
\end{equation}
\begin{equation} 
\label{GRU_hidden}
y_{t+1} = h_{t+1}
\end{equation}
\begin{figure}
	\includegraphics[width=\textwidth]{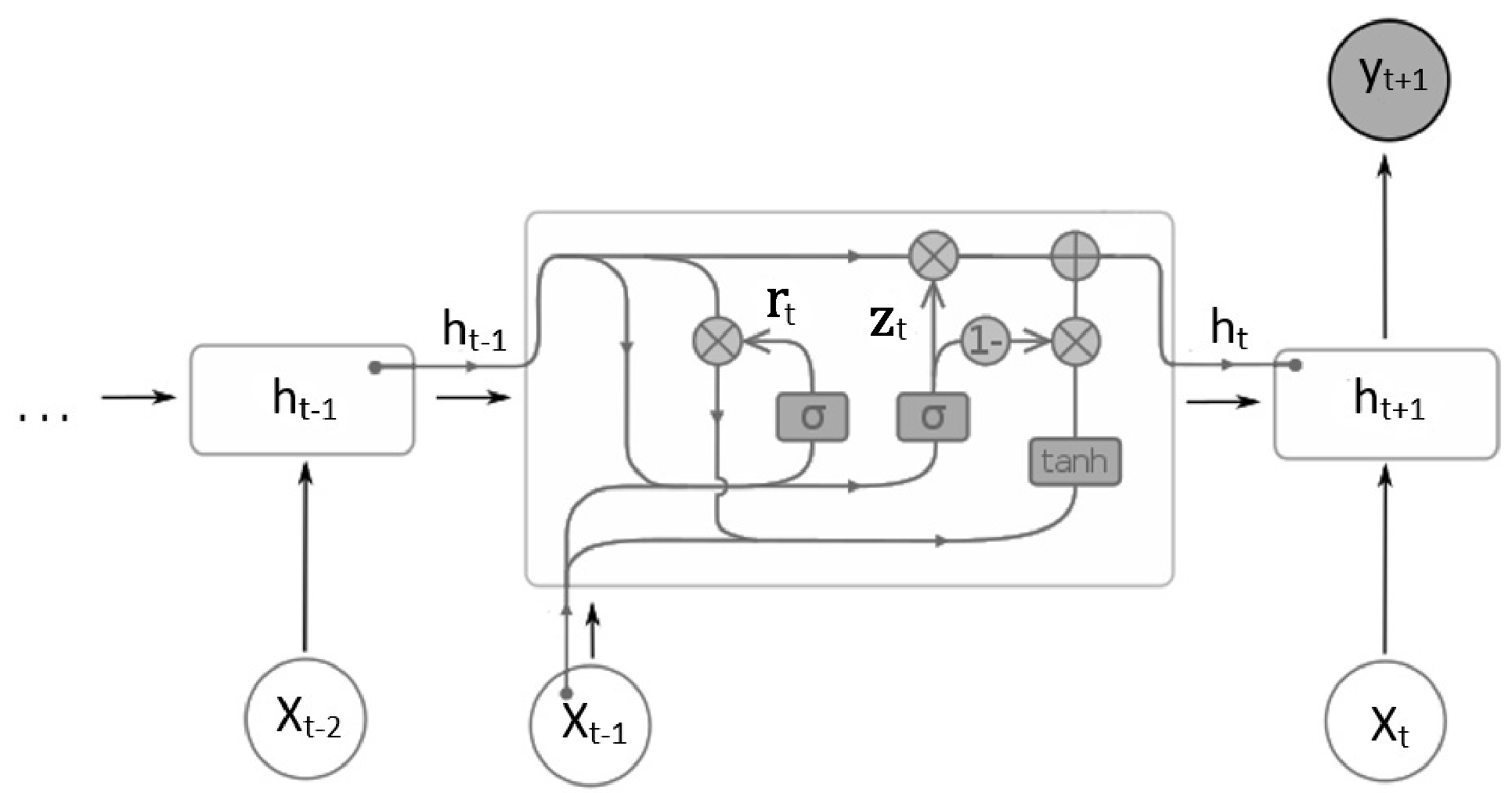}
	\caption{An Unfolded GRU Cell}
	\label{fig:GRU}
\end{figure}
In the VAR model proposed by Lütkepohl \cite{lutkepohl2005new}, as illustrated in Equation (\ref{var}), the estimated value is represented as a linear function of its preceding values within the last $t$ time-intervals. Additionally, this model incorporates a constant vector $c$ and an unobservable zero-mean white noise vector $e_{t'}$. 
In the proposed model, as illustrated in Fig. \ref{fig:PredictionModel} and described by Eq. (\ref{MyModel}), three recurrent neural network layers (denoted as $rnn\_layer$) are employed to perform nonlinear supervised feature extraction on the input vectors ($x$) from Eq. (\ref{features}) corresponding to the last $t$ time-intervals. Subsequently, significant features ($f$) are extracted. Ultimately, the final layer of the model, which utilizes an identity activation function akin to linear regression as per Eq. (\ref{MyModel2}), estimates the coefficient vector for the subsequent time-interval based on the extracted features.
In the feature extraction component of the model, each recurrent cell can be represented by the previously described cells. By concatenating input features with the output of the preceding cell, the issue of vanishing gradients can be mitigated, thereby accelerating the learning process \cite{he2016deep}.
\begin{equation} 
\label{var}
y_{t'} = c + A_{t'-1}y_{t'-1} + ... + A_{t'-t}y_{t'-t} + e_{t'}
\end{equation}
\begin{equation} 
\label{features}
x = coe\_vectors[t'-t\;\;to\;\;t'-1]
\end{equation}
\begin{equation} 
\label{MyModel}
f = rnn\_layer([rnn\_layer([rnn\_layer(x),x]),x],x])
\end{equation}
\begin{equation} 
\label{MyModel2}
y_{t'} = W.f + b
\end{equation}
In the proposed model, the training process takes into account the Mean Squared Error (MSE) cost function, as shown in Equation (\ref{MSE_cost}), over a period of $(T)$ time-intervals. The Root Mean Square Propagation (RMS-Prop) optimizer is employed to minimize this cost function effectively.

\begin{equation} 
\label{MSE_cost}
MSE = \frac{\sum_{t = 1}^{T}(\hat{y}_{t}-y_{t})^2}{T}
\end{equation}

\section{Experiments}
To train and evaluate our proposed model, we utilized a dataset comprising 45 consecutive days of urban travel data from Tapsi, a top online taxi-hailing service provider in the Middle East\footnote{It is important to note that the dataset used in this study is not openly available to other researchers, as it was shared with the authors due to a connection with the company's CTO.}. The dataset consists of 100 samples, each containing 45 consecutive days from different periods in 2018 and 2019, showcasing a variety of weather-conditions and holidays. In each sample, 38 days are allocated for training and clustering, while the remaining 7 days are used for classification and model validation. The testing data classification and model validation rely on the clusters derived from the K-means algorithm applied to the training data in a four-dimensional space, as previously mentioned. The model's performance is assessed by reporting the mean of the errors across these 100 training-test pairs.

\subsection{Four-Dimensional K-Means Clustering}
\label{four_dimensional_kmeans_clustering}
\begin{figure}
	\includegraphics[width=0.6\textwidth]{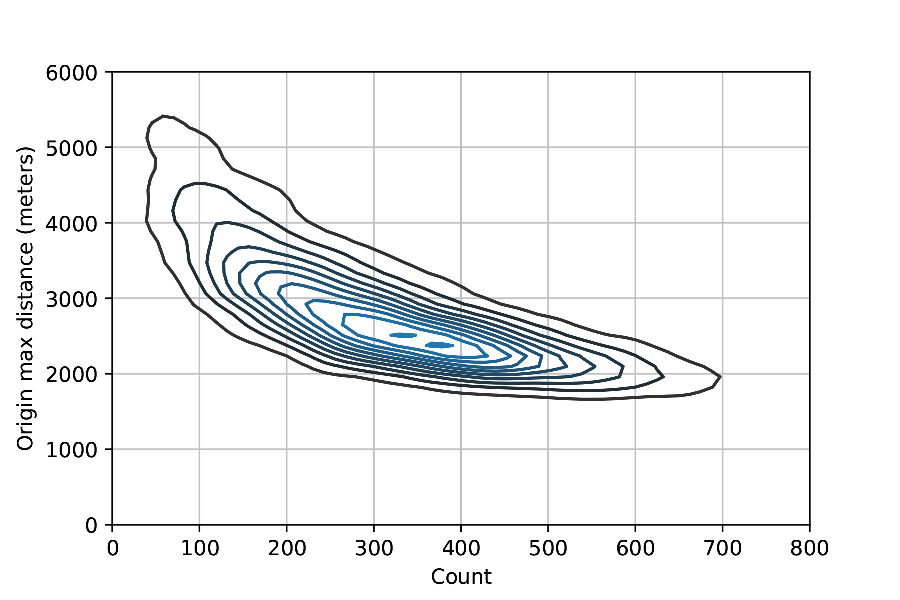}
	\caption{The maximum intra-cluster distance within the origin zones and the corresponding travel frequency count.}
	\label{fig:Origin_max_dist}
\end{figure}

\begin{figure}
	\includegraphics[width=0.6\textwidth]{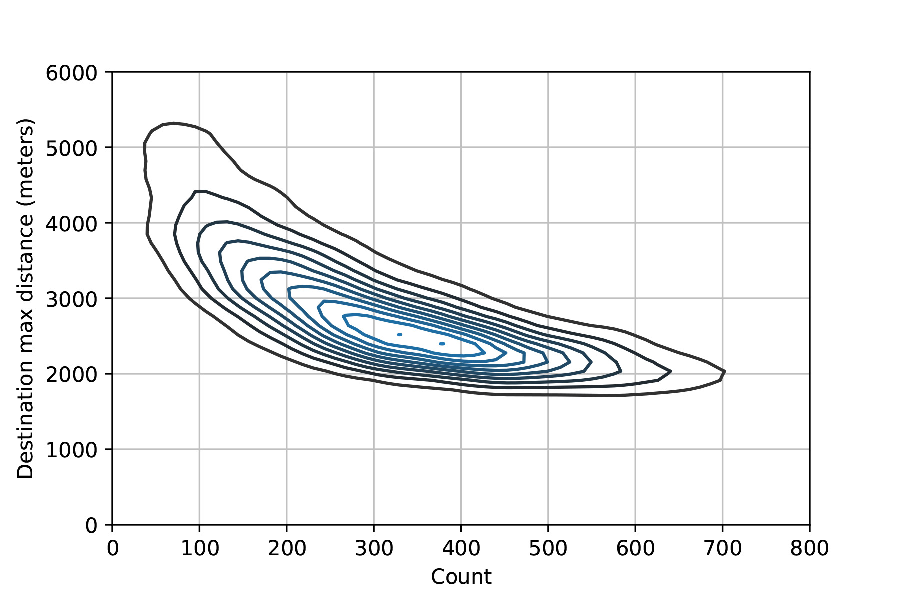}
	\caption{The maximum intra-distance within the destination zones of clusters and the corresponding number of travel occurrences.}
	\label{fig:Destination_max_dist}
\end{figure}

Initially, the K-means clustering algorithm is employed in a four-dimensional space to identify significant OD flows. To incorporate the maximum intra-distance constraint within origin and destination zones, the K-means algorithm is executed for varying numbers of clusters. It is important to note that an increase in the number of clusters leads to a wider confidence interval for the predictions, potentially resulting in less informative outcomes.
For smaller travel cluster sizes, the prediction process becomes increasingly challenging. In essence, these travel clusters exhibit reduced predictability and increased randomness. A confidence interval deemed acceptable by domain experts from the data provider company consists of 10,000 travel clusters, each with an average maximum intra-distance of 2,900 meters for both origin and destination zones. It is important to note that the limitation on zones' maximum intra-distance is contingent upon the maturity of the travel data.
In this study, clusters deemed insignificant, specifically those with fewer than 10 journeys within a five-week period, were excluded from the analysis. As demonstrated in Fig. \ref{fig:Origin_max_dist} and Fig. \ref{fig:Destination_max_dist}, an increase in the maximum intra-distance of origin and destination zones within each cluster leads to an improvement in travel data quality or an increase in the number of recorded journeys, and vice versa. It is worth noting that small origin or destination zones with high travel frequencies may encompass hotspots such as shopping centers, universities, and other similar establishments.
The outcomes of the three previously mentioned clustering techniques are presented in Table \ref{tab:clustering}. Clusters containing less than 10 trips are deemed insignificant in terms of travel pattern analysis. The four-dimensional K-means clustering approach is capable of extracting meaningful travel patterns due to the incorporation of both origin and destination points of the trips. Based on the clustering results obtained from our dataset, this method has generated fewer travel clusters compared to other existing methodologies.

\begin{table}
	\caption{Comparative Analysis of Three Distinct Clustering Techniques for Travel Data}\label{tab:clustering}
	\begin{tabular}{lllll}
			\hline\noalign{\smallskip}
			Clustering & OD size & Zones size avg  & \#Clusters with &\#Clusters\\
			methods& &(meters)&more than 10 travels&\\
			\noalign{\smallskip}\hline\noalign{\smallskip}
			Grid & 30976 (176*176) & = 3000 & 15984 & 20291 \\ 
			2D K-means & 25600 (160*160) &$\approx$ 3000 & 24998 & 25595 \\ 
			4D K-means & 10000 (vector) &$\approx$ 2900 & 9973 & 10000 \\
			\noalign{\smallskip}\hline
	\end{tabular}
\end{table}

\subsection{Prediction Model}
In Section \ref{prediction_model_section}, we employ the MSE as the cost function for the model during the training phase. This metric imposes a higher penalty on larger errors compared to the Mean Absolute Error (MAE) as shown in Eq. (\ref{MAE_cost}). In this section, we utilize both MAE and MSE metrics to compare the performance of different models and examine various aspects of their performance. Additionally, we introduce another metric called Mean Absolute Percentage Error (MAPE@1) as presented in Eq. (\ref{MAPE@1_cost}). This metric expresses errors as a percentage, which is independent of the problem context, thereby providing a more comprehensible representation of the models' performance.

The study employs hourly time-interval data to predict the number of travels for each hour in the clusters. The size of the time-intervals and travel clusters may impact the prediction confidence interval. K-means clustering is executed, and significant OD flows are extracted. To minimize the number of trainable variables in the model, the NMF algorithm is utilized to reduce the OD vector dimension from 10000 to 100.

The results of dimension reduction using NMF yielded a mean a MSE of 0.016 and a MAE of 0.079 on the training dataset. Furthermore, Table \ref{tab:NMF} demonstrates that the NMF technique has a substantial influence on the number of trainable parameters for the three proposed prediction models, each utilizing distinct recurrent layers.
\begin{equation} 
\label{MAE_cost}
MAE = \frac{\sum_{t = 1}^{T}|\hat{y}_{t}-y_{t}|}{T}
\end{equation}

\begin{table}
	\caption{The number of trainable variables for three different models, both with and without the application of the NMF algorithm.}\label{tab:NMF}
	\begin{tabular}{lll}
			\hline\noalign{\smallskip}
			Model & With NMF & Without NMF \\
			\noalign{\smallskip}\hline\noalign{\smallskip}
			Stacked\_RNN & 10,619,600 & 59,900,481,500 \\ 
			Stacked\_GRU & 31,456,600 & 176,691,224,500 \\ 
			Stacked\_LSTM & 41,875,100 & 235,086,596,000 \\ 
			\noalign{\smallskip}\hline
	\end{tabular}
\end{table}

\begin{figure}
	\includegraphics[width=0.9\textwidth]{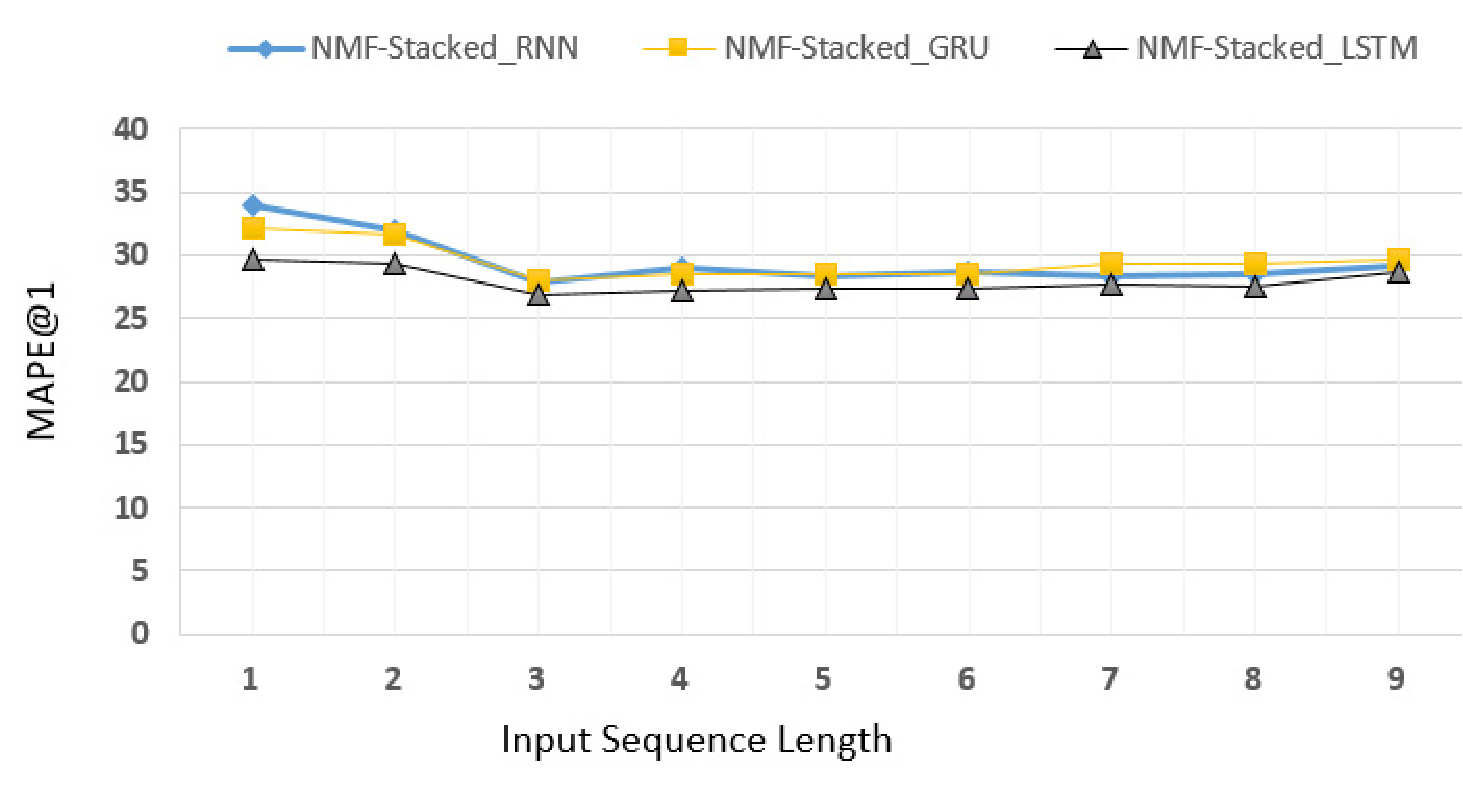}
	\caption{The impact of varying input time lengths on the performance of the proposed models.}
	\label{fig:Time_Length}
\end{figure}

\begin{figure}
	\includegraphics[width=1\textwidth]{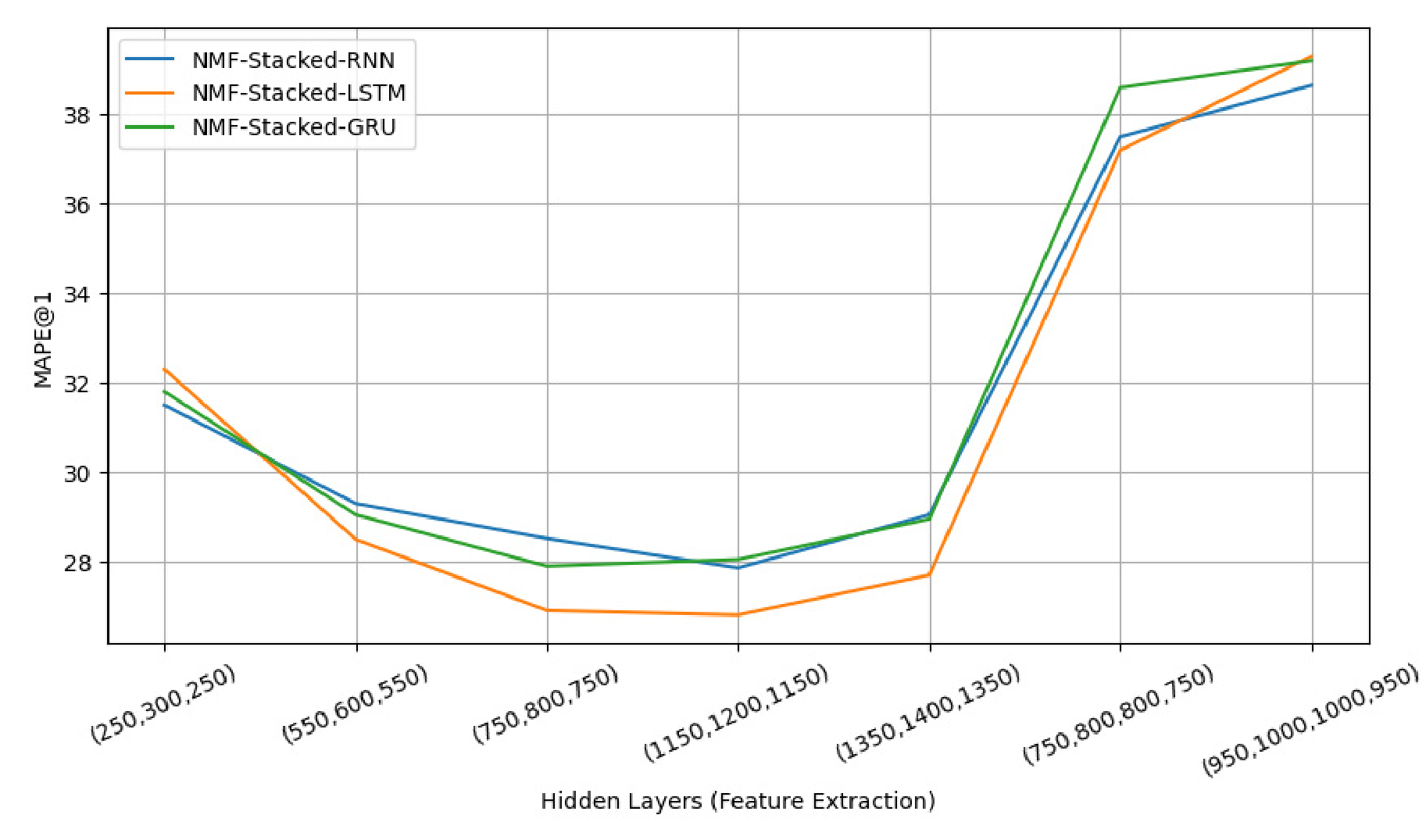}
	\caption{Comparison of various hidden layer architectures with respect to the MAPE at 1 on the test data.}
	\label{fig:finding_best_parameters}
\end{figure}

In order to determine the optimal input time length, nine distinct input time lengths were examined for the proposed model. The MAPE@1 was employed to assess the model's error in predicting non-zero demands.
Three different recurrent cell types, as introduced earlier, were utilized in the recurrent layers of the model during this evaluation. Furthermore, the MSE, MAE, and MAPE@1 metrics were computed on denormalized data across all tests.
As illustrated in Figure \ref{fig:Time_Length}, a time series sequence of three hours is considered for the proposed model, which entails forecasting based on the coefficient vectors from the preceding three hours, time features, and weather-conditions. In order to reduce the prediction error of the model, weather forecast data can be incorporated at the final layer, concatenated with the output of the supervised feature extraction.
It appears that extending the length of sequences does not contribute to enhancing the prediction accuracy; rather, it leads to increased computational time. Furthermore, as illustrated in Fig. \ref{fig:finding_best_parameters}, various layer sizes are examined to determine the optimal architecture for hidden layers (pertaining to the feature extraction component), and the most suitable configuration is subsequently chosen. 

\begin{equation} 
\label{MAPE@1_cost}
MAPE@1 = \frac{1}{\#(y_t \ge 1)}\sum_{y_t \ge 1}\frac{|y_t - \hat{y}_t|}{y_t} * 100\%
\end{equation}

\begin{figure}
	\includegraphics[width=0.95\textwidth]{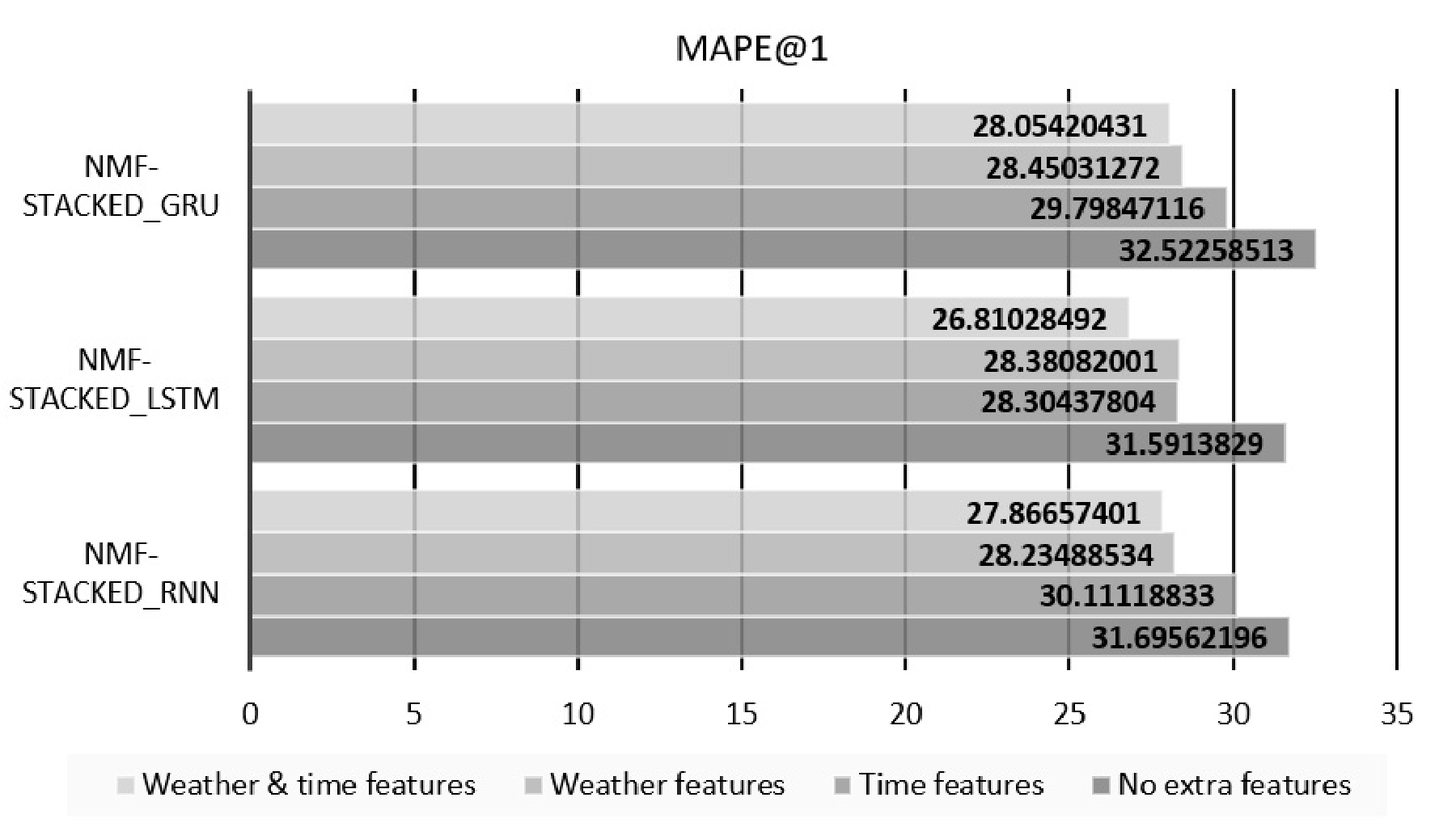}
	\caption{Influence of Additional Features on the Performance of the Proposed Models}
	\label{fig:Features}
\end{figure}

As elucidated in Section \ref{prediction_model_section}, the proposed model incorporates a supervised feature extraction component, thereby eliminating the need for manual feature selection. To assess the influence of time and weather features on the model's prediction capabilities, we conducted training and testing with and without these additional features. Figure \ref{fig:Features} demonstrates that the inclusion of weather features results in a more significant reduction in the MAPE@1 metric compared to time features, indicating their greater effectiveness.

To provide more practical insights, we further analyzed the prediction results under different conditions, such as various days of the week, holidays, and specific weather-conditions. Our analysis revealed that the model's performance varies across different days of the week, with higher prediction accuracy observed during weekdays compared to weekends. This can be attributed to more regular and predictable travel patterns during weekdays, as people tend to follow routine schedules for work or school. On weekends and holidays, travel patterns become more diverse and less predictable, leading to a slight decrease in prediction accuracy.

Furthermore, we observed that the model's performance is affected by extreme weather-conditions, such as heavy rain or snow, which can cause significant disruptions in travel patterns. In such cases, incorporating real-time weather data and adjusting the model accordingly can help improve prediction accuracy and provide more reliable information for the online taxi-hailing system.

The proposed model can benefit from incorporating real-time weather data and considering the variations in travel patterns across different days of the week and holidays. These insights can help online taxi-hailing systems to better allocate resources, adjust pricing strategies, and ultimately enhance their overall efficiency and user satisfaction.

In order to evaluate the prediction accuracy of our proposed model in comparison to existing models, we implemented several of them. The most basic among these is referred to as the calendar model, which can be categorized into three distinct types: 1) Hourly calendar model, which computes the mean travel count for the previous three hours, 2) Daily calendar model, which calculates the mean travel count for the preceding seven days, and 3) Weekly calendar model, which determines the mean travel count for the past five weeks, specifically for the given hour and day.
The second model implemented for comparison is the MLP, which shares a similar structure with our proposed model, except that the recurrent layers are substituted by simple neuron layers. The input data for the MLP model closely resembles that of our proposed model, with the exception of the time series axis. Consequently, all features from the previous three hours are incorporated into a single vector.
Additionally, we employ the NMF algorithm on the input of the implemented models, which is indicated by the prefix 'NMF' in their names. Other models that utilize input data akin to the MLP model include SVM \cite{smola2004tutorial} with Radial Basis Function (RBF) kernel, K-Nearest Neighbors (KNN) \cite{altman1992introduction} with uniform distance weight, and Random Forest \cite{breiman2001random}.
Moreover, we implemented the NMF-VAR \cite{li2018hybrid}, as described earlier, which has an input data structure similar to our proposed model. With the exception of the weekly and daily calendar models, the other models leverage features from the previous three hours to predict travel in the subsequent hour.
\begin{figure}
	\includegraphics[width=0.9\textwidth]{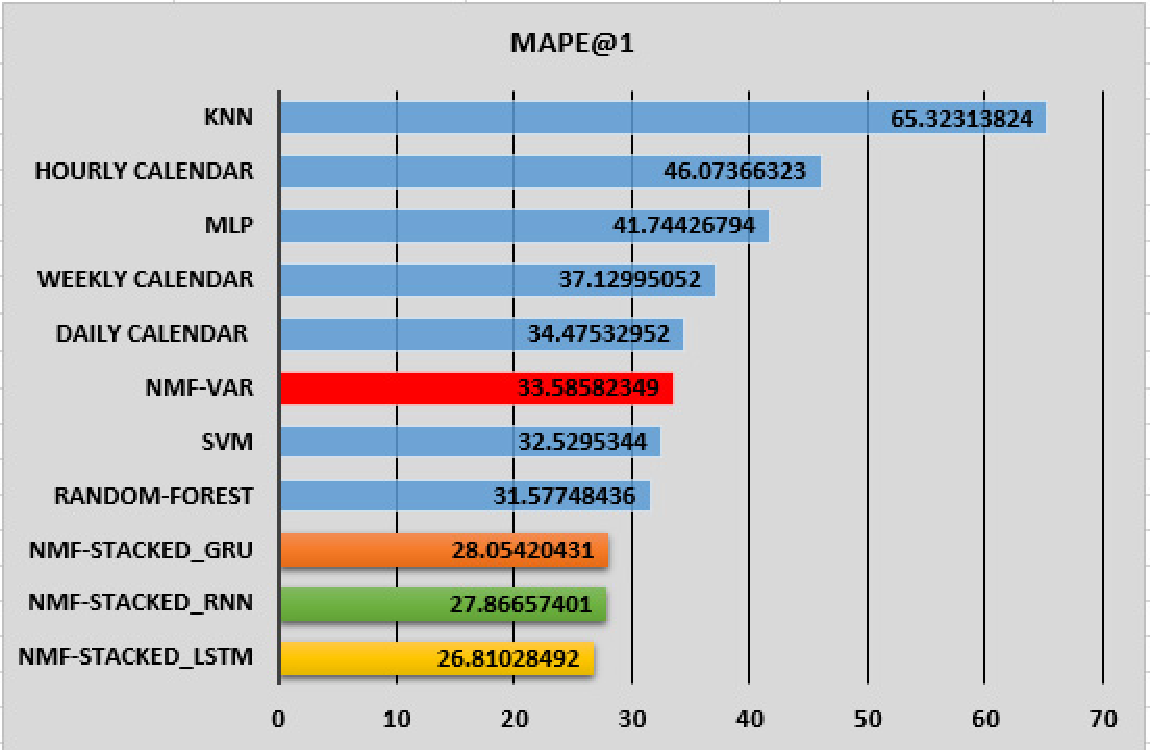}
	\caption{Comparison of models based on the MAPE@1 metric.}
	\label{fig:Models_MAPE1}
\end{figure} 
\begin{figure}
	\includegraphics[width=0.9\textwidth]{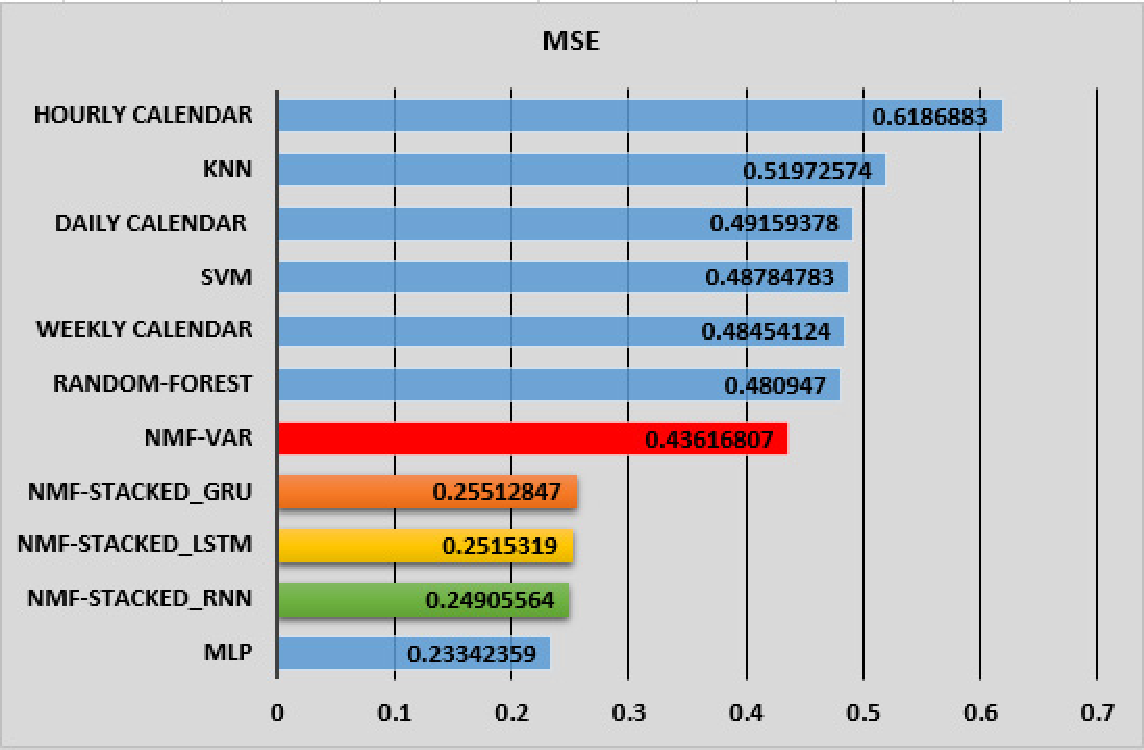}
	\caption{Comparison of models based on the MSE metric.}
	\label{fig:Models_MSE}
\end{figure}

\begin{figure}
	\includegraphics[width=0.9\textwidth]{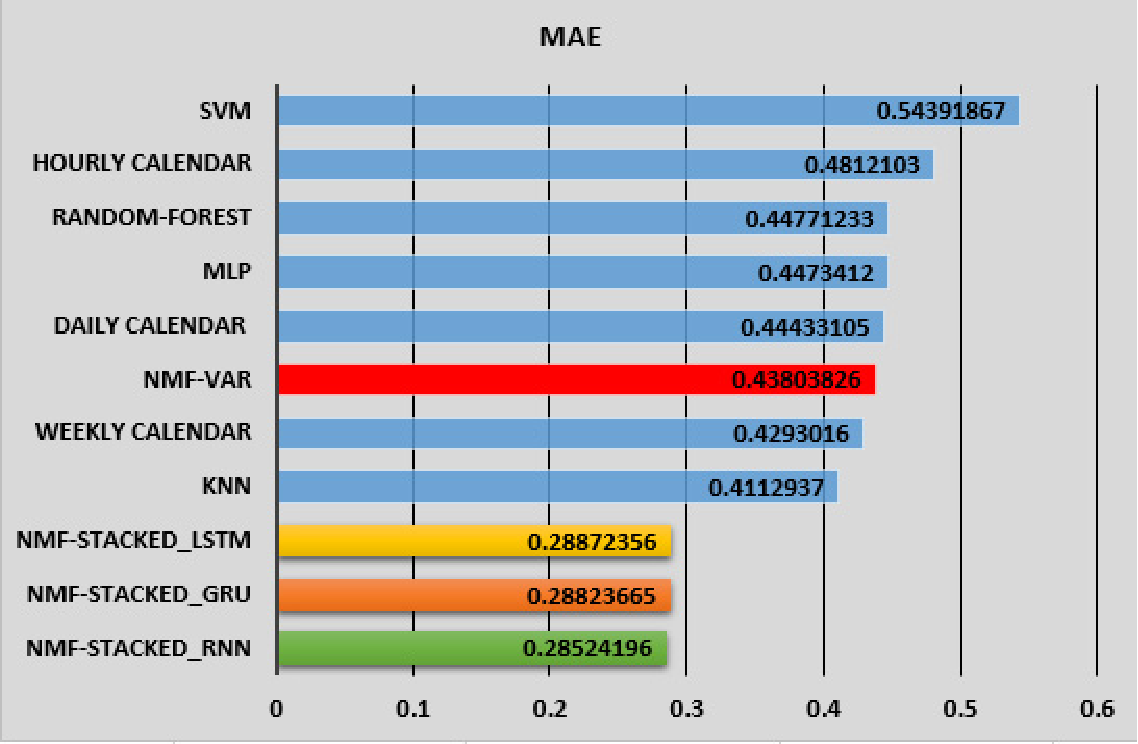}
	\caption{Comparison of models based on the MAE metric.}
	\label{fig:Models_MAE}
\end{figure} 

\begin{figure}
	\includegraphics[width=0.9\textwidth]{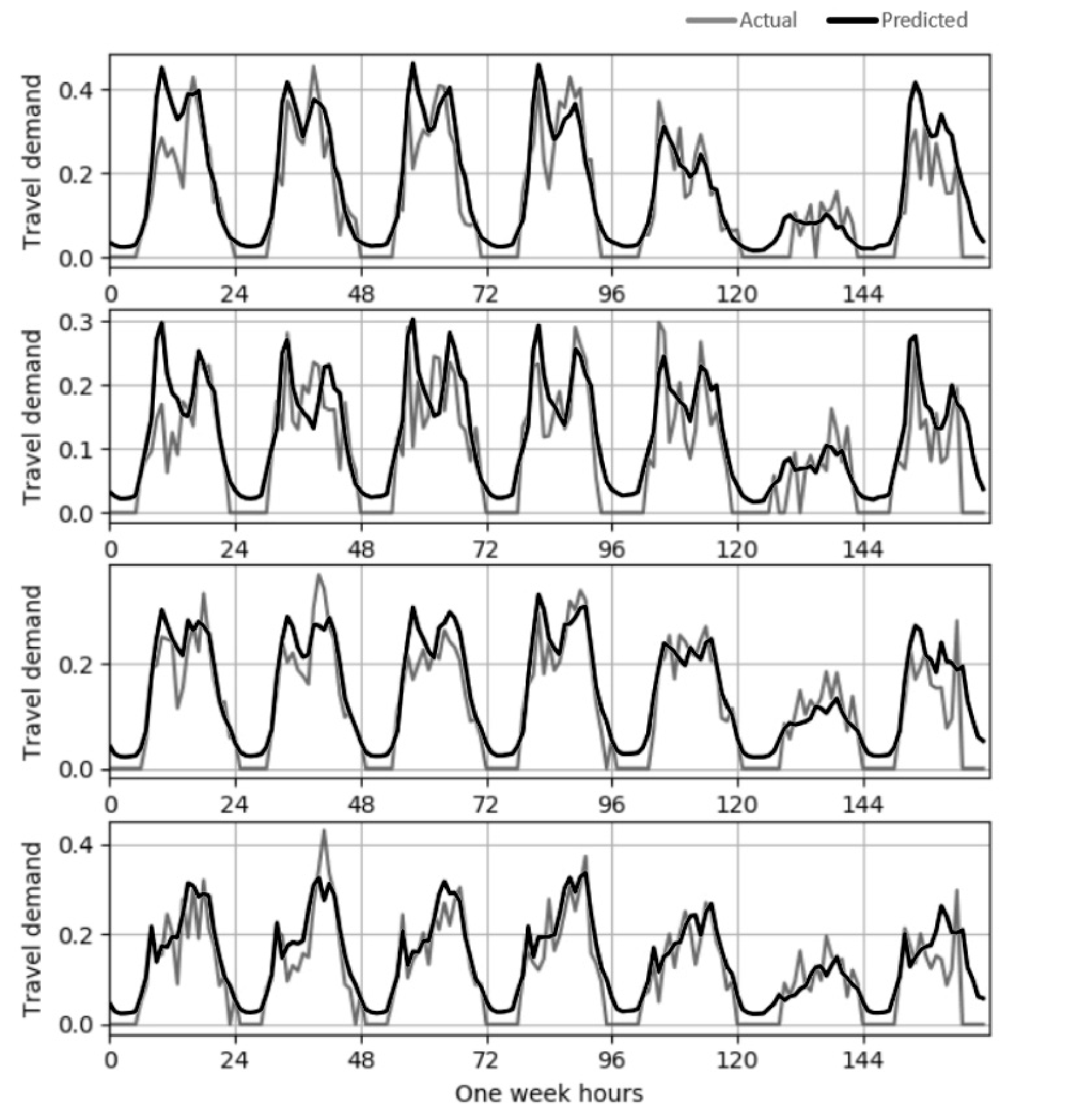}
	\caption{The NMF-Stacked LSTM model's prediction results were obtained for four randomly selected clusters out of the 10,000 total clusters.}
	\label{fig:LSTM_Prediction}
\end{figure}

As depicted in the bar chart in Fig. \ref{fig:Models_MSE}, it appears that the prediction accuracy of the MLP model surpasses that of the other models. However, based on the bar charts in Fig. \ref{fig:Models_MAE} and Fig. \ref{fig:Models_MAPE1}, the MLP model does not predict nonzero travel counts as effectively as the recurrent models. In essence, taking all the bar charts into account, the three proposed models demonstrate superior performance compared to the existing models.
As previously discussed, the stacked recurrent model is capable of extracting non-linear relationships among the feature vectors from the past three hours (weather-conditions, time, and travel count) and the coefficient vector for the subsequent hour. In contrast, the NMF-VAR model only determines the coefficients for a linear model based on the travel count from the previous three hours. The three proposed stacked recurrent models exhibit similar predictability; however, according to Fig. \ref{fig:Models_MAPE1}, it appears that the higher complexity of the NMF-Stacked LSTM model enables it to extract more valuable features during the feature extraction phase of linear regression, resulting in a more accurate prediction for non-zero values.
The predicted coefficient vectors were transformed into travel vectors by multiplying them with the basis matrix. In Fig. \ref{fig:LSTM_Prediction}, four out of 10,000 travel clusters are randomly selected to compare the travel prediction results of the NMF-Stacked LSTM model with the actual values.
We conducted the same comparisons for NMF-VAR, NMF-Stacked RNN, NMF-Stacked GRU, and NMF-Stacked LSTM using travel data with 30-minute time-intervals. Reducing the size of the confidence interval adversely affects prediction accuracy; however, the results obtained are more practical. The models were trained based on features from the previous 180 minutes to predict the travel count for the upcoming 30 minutes.
\begin{figure}
	\includegraphics[width=0.7\textwidth]{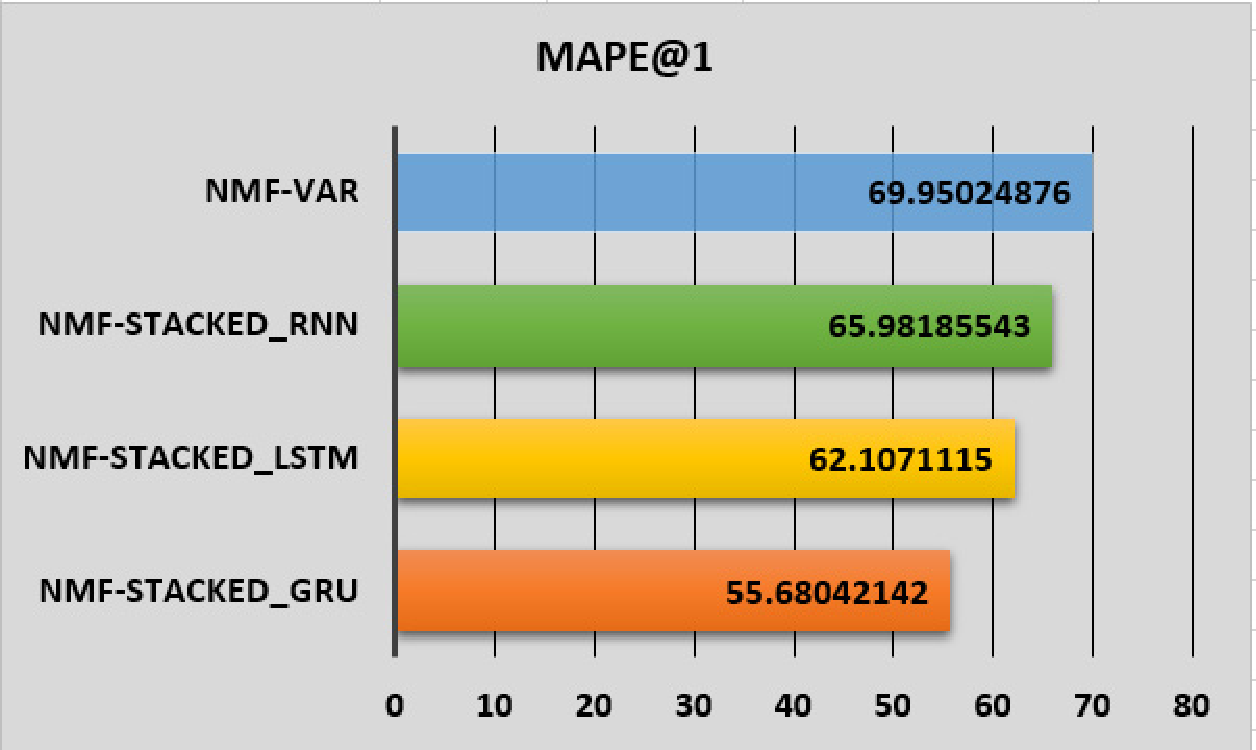}
	\caption{Comparison of models based on the MAPE@1 metric for a narrower time-interval.}
	\label{fig:Models_MAPE1_30_min}
\end{figure} 
\begin{figure}
	\includegraphics[width=0.7\textwidth]{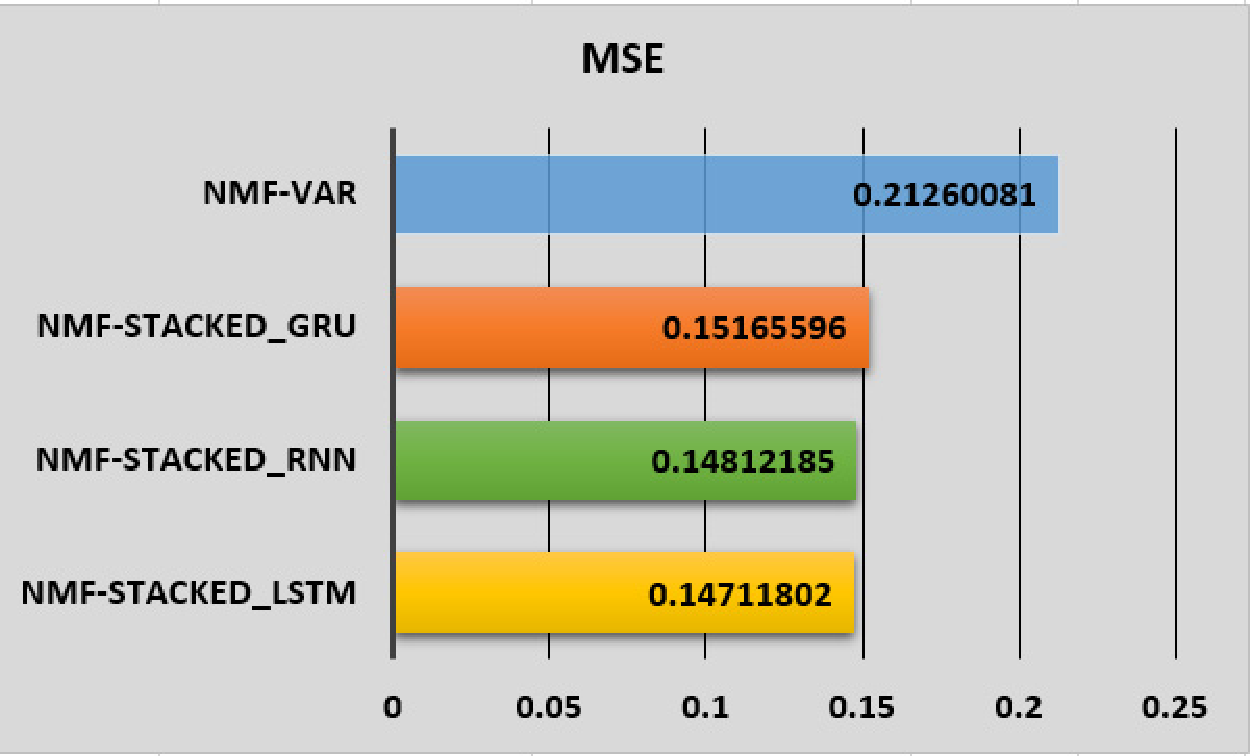}
	\caption{Comparison of models based on the MSE metric for a narrower time-interval.}
	\label{fig:Models_MSE_30_min}
\end{figure} 
\begin{figure}
	\includegraphics[width=0.7\textwidth]{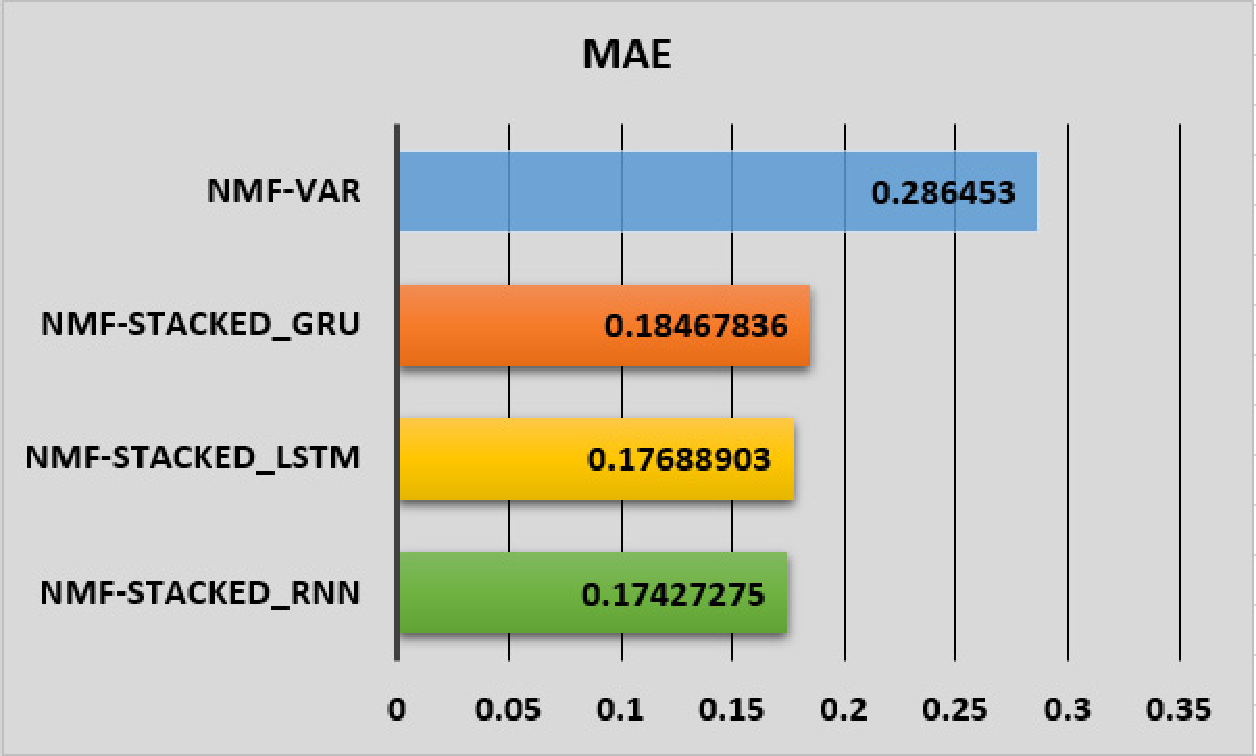}
	\caption{Comparison of models based on the MAE metric for a narrower time-interval.}
	\label{fig:Models_MAE_30_min}
\end{figure} 
As demonstrated in Fig. \ref{fig:Models_MSE_30_min} and Fig. \ref{fig:Models_MAE_30_min}, the MSE and MAE values have decreased compared to those in Fig. \ref{fig:Models_MSE} and Fig. \ref{fig:Models_MAE}. This reduction can be attributed to the decrease in travel count within narrower time-intervals. According to Fig. \ref{fig:Models_MAPE1_30_min}, the MAPE@1 value has increased, as predicting for tighter intervals is more challenging, as previously explained. The results indicate that the NMF-Stacked GRU model can predict nonzero travel counts with significantly better accuracy than the other three models. However, based on the MAE and MSE metrics, the predictability of the proposed recurrent models is relatively similar to one another.

\section{Conclusion}
Utilizing online taxi-hailing passenger travel data, we developed a prediction model. Initially, we employed the K-means clustering algorithm in four-dimensional space to group meaningful travel flows into 10,000 clusters. We demonstrated that these clusters exhibit superior quality compared to Grid-based and 2D K-means clustering results. Following the clustering of travel data, we aggregated the demand values for each travel cluster in hourly time-intervals.
To improve prediction accuracy, weather-conditions and time features were integrated into the input vector. These supplementary features led to an approximate 5\% enhancement in MAPE@1 for non-zero travel count prediction. Additionally, we proposed a neural network model consisting of a stacked recurrent section for supervised feature extraction and a linear regression section. The feature extraction segment has the ability to discern non-linear relationships among the features.
Based on our experimental results, the proposed model demonstrated superior performance compared to other popular models, achieving approximately 5-7\% improvement in MAPE@1, 0.19 in MSE, and 0.13 in MAE. Simple-RNN, LSTM, and GRU, as recurrent neural network cells within the feature extraction component of the model, yielded similar outcomes. However, for 30-minute time-interval predictions, the proposed model with the GRU recurrent cell exhibited approximately 14\%, 10\%, and 7\% better MAPE@1 compared to the NMF-VAR, NMF-Stacked RNN, and NMF-Stacked LSTM models, respectively.
The diminished prediction error could assist taxi-hailing companies in implementing a more equitable pricing system and achieving a better balance between supply and demand. Furthermore, employing such a prediction model would enable ride-sharing services to activate their offerings when the likelihood of matching is high, thereby reducing costs. In future research, additional external features, such as hotspot categories and travel prices, will be considered to improve the accuracy of OD flow prediction. Another potential challenge is to decrease the size of origin and destination hotspot zones while preserving prediction accuracy.


%
%

\bibliographystyle{spbasic}      
\bibliography{paper}   

\end{document}